\documentclass[10pt]{article}
\usepackage[letterpaper,margin=1in]{geometry}

\usepackage[sc]{mathpazo}
\usepackage{microtype}

\usepackage{amsmath,amssymb,amsfonts}
\numberwithin{equation}{section}

\usepackage{graphicx}
\usepackage{booktabs}
\usepackage{xcolor}
\definecolor{rspaBlue}{RGB}{0,114,178}

\usepackage{siunitx}
\usepackage{algorithm}
\usepackage{algorithmic}

\usepackage[acronym]{glossaries}
\makeglossaries

\usepackage{hyperref}
\usepackage[capitalize,noabbrev]{cleveref}

\newacronym[plural=FEAs]{fea}{FEA}{Finite Element Analysis}
\newacronym{sindy}{SINDy}{Sparse Identification of Nonlinear Dynamics}
\newacronym{wls}{WLS}{Weighted Least-Squares}
\newacronym{gls}{GLS}{Generalized Least-Squares}
\newacronym{blue}{BLUE}{Best Linear Unbiased Estimator}
\newacronym{stls}{STLS}{Sequential Threshold Least Squares}
\newacronym{ode}{ODE}{Ordinary Differential Equation}
\newacronym{pde}{PDE}{Partial Differential Equation}
\newacronym{stridge}{STRidge}{Sequential Thresholded Ridge Regression}
\newacronym{sr3}{SR3}{Sparse Relaxed Regularized Regression}
\newacronym{gp}{GP}{Gaussian process}
\newacronym{pce}{PCE}{polynomial chaos expansion}

\title{\bfseries Multi-Fidelity SINDy: Sparse Discovery of Nonlinear Dynamical Systems with Fidelity-Weighted Measurements}

\author{
\begin{tabular}{c}
Filippo Zacchei$^{1,*}$, Ana Larra\~{n}aga$^{2}$, Attilio Frangi$^{3}$,\\
Andrea Manzoni$^{1}$ and Steven~L. Brunton$^{2}$\\[0.6em]
\small
\begin{tabular}{p{0.92\textwidth}}
\centering $^{1}$MOX -- Department of Mathematics, Politecnico di Milano, Piazza Leonardo da Vinci 32, 20133 Milano, Italy\\
\centering $^{2}$Department of Mechanical Engineering, University of Washington, Seattle, Washington 98195, United States\\
\centering $^{3}$Department of Civil and Environmental Engineering, Politecnico di Milano, Piazza Leonardo da Vinci 32, 20133 Milano, Italy\\[0.4em]
\centering $^{*}$corresponding author: \texttt{filippo.zacchei@polimi.it}
\end{tabular}
\end{tabular}
}

\date{}

\begin{document}

\maketitle

\begin{abstract}
Data from simulations and experiments are rarely noise-free and often exhibit heterogeneous levels of fidelity. Measurement uncertainty may vary across repeated observations, sensing devices, or even within a single experiment. This work addresses the problem of discovering nonlinear dynamical systems from such inhomogeneous data. We extend the Sparse Identification of Nonlinear Dynamical Systems (SINDy) framework to account for variable noise levels by combining Ensemble SINDy and Weak SINDy within a weighted regression formulation derived from generalized least squares. A statistical justification for the weighting strategy is also provided. The methodology is validated on several benchmark systems, including ordinary and partial differential equations. In addition, we show the benefit of multi-fidelity integration for forecasting the dynamics of a double pendulum system. The results confirm that the proposed approach mitigates the adverse effects of heteroscedastic noise and that repeated, low-cost, low-quality measurements can improve model recovery, in some cases matching or outperforming reconstructions obtained using only high-fidelity data.
\end{abstract}

\noindent\textbf{Keywords:} nonlinear dynamics, sparse regression, model discovery, uncertainty quantification, probabilistic forecasting, multi-fidelity

\section{Introduction}

In recent decades, engineering has undergone a significant revolution. Advances in computational modeling have shifted the field from a primary reliance on physical experimentation toward an increasingly virtual paradigm \cite{chandrupatla2021,brunton2019}. High-fidelity simulations now enable accurate representations of complex dynamical systems and have supported the development of digital twins \cite{tao2024} for prediction \cite{kapteyn2021}, monitoring \cite{torzoni2024}, and control \cite{mcclellan2022}. However, such simulations are often computationally expensive, particularly when relying on finite element solvers or when dealing with  multiphysics settings, which limits their use in real-time applications and in tasks requiring repeated evaluations, such as uncertainty quantification \cite{quarteroni2016,SEELINGER2025113542}.

To address these limitations, surrogate models have emerged as efficient approximations of high-fidelity simulators \cite{forrester2008,benner2015,ZACCHEI2024104902}. Among them, data-driven surrogates are especially attractive because they learn system behavior directly from observations \cite{brunton2019,samadian2025}. In the context of dynamical systems, the \gls{sindy} framework \cite{brunton2016} is appealing because it identifies governing equations directly from measured trajectories while promoting parsimonious model structure. The resulting models are interpretable and often more robust to out-of-distribution extrapolation than black-box alternatives such as Gaussian processes or neural networks. These properties make \gls{sindy} especially attractive for digital-twin applications \cite{brunton2020,willard2022integratingscientificknowledgemachine}.

The performance of data-driven surrogate models, however, depends on the possibility to access sufficiently rich high-fidelity datasets obtained from either physical experiments or detailed numerical simulations. Such datasets can be costly or impractical to acquire, especially in experimental settings producing noisy measurements and limited sensing capability, or in virtual settings that require long and expensive simulations \cite{willard2022integratingscientificknowledgemachine}. By contrast, lower-fidelity data are often much easier to obtain, whether from simplified numerical models or from cheaper and less precise sensing devices. A natural objective is therefore to exploit such low-fidelity information while preserving, as much as possible, the predictive accuracy associated with models trained on high-fidelity data \cite{peherstorfer2018}. 
This objective poses several challenges for system identification methods, such as \gls{sindy}. Low-fidelity data are often heavily contaminated by noise, which can severely degrade the estimation of time derivatives, a key step in dynamical system identification. As a result, the regression stage in standard \gls{sindy} may become unreliable \cite{kaiser2018}. Recent advances, such as ensemble \gls{sindy} \cite{fasel2022} and weak \gls{sindy} \cite{messenger2021_1,messenger2021_2}, allow for an improved robustness to noise and the use of lower-quality data. However, extending \gls{sindy} to multi-fidelity settings remains a challenging task, since existing formulations do not provide a principled mechanism for weighting and combining data sources of different reliability while preserving the information carried by high-fidelity observations.

In this work, we introduce a multi-fidelity training strategy for \gls{sindy} that accounts for heterogeneous measurement reliability. Building on weak \gls{sindy} and ensemble \gls{sindy}, and incorporating a covariance-aware weighting of the sparse regression step, the proposed approach enables accurate and interpretable model identification when the data arise from heterogeneous sources or exhibit nonuniform noise levels within a single trajectory.

Multi-fidelity data-driven surrogate modeling provides strategies for learning models from data collected at different fidelities, improving the accuracy of models when high fidelity data are limited \cite{godino,peherstorfer2018}. Foundational works in this area concern \gls{gp} regression, where cross-fidelity dependence is modeled through co-kriging and auto-regressive formulations \cite{le2014recursive, forrester2007multi, perdikaris1, Peridkaris2, raissi2016deepmultifidelitygaussianprocesses}.
Alternative formulations are based on \gls{pce} and introduce structured corrections across fidelities (e.g., additive or multiplicative) within a single expansion \cite{BRYSON2017121}, and multi-level or sparse multi-fidelity \gls{pce} variants exploit hierarchical fidelity organization while controlling complexity through sparsity \cite{CHENG2019360, piazzola2023comparing, KENT2026114761}.
Hybrid methods bridge these paradigms, using schemes that combine polynomial expansions with kriging-style correlation modeling \cite{Du}. 
More recently, scalable machine-learning surrogates have been developed to learn nonlinear fidelity mappings directly from joint low-/high-fidelity datasets \cite{zacchei2026multi, GUO2022114378, conti2026progressivemultifidelitylearningneural, villatoro}, including composite-network and multi-fidelity PINN frameworks \cite{MENG2020109020, PENWARDEN2022110844, regazzoni2021physics}, neural operators \cite{demo2023deeponet,HOWARD2023112462,rowbottom2025multi}, residual networks \cite{davis}, and probabilistic neural-process formulations for multi-fidelity regression \cite{niu24d}. A recent work also introduces an approximate control variate framework to define new multifidelity Monte Carlo estimators for linear regression models \cite{qian2024multifidelity}. Application-driven studies illustrate these ideas in data-fusion settings and in hybrid correction pipelines that combine reduced-order low-fidelity solvers with learned high-fidelity adjustments \cite{e22091022, SAJJADINIA2022105699}.

Within this landscape, multi-fidelity \gls{sindy} offers a distinct perspective when the data consist of time-resolved trajectories and the objective is explicit governing-equation discovery rather than black-box input--output emulation. In particular, such a formulation is well suited to settings in which: (i) fidelity differences manifest primarily as heteroscedastic measurement noise across sensors, experiments, or simulations; (ii) many low-fidelity trajectories are available, but only a limited number of high-fidelity trajectories can be used for coefficient refinement; and (iii) interpretability, mechanistic insight, or downstream tasks such as control and extrapolation under structural constraints are central. In these settings, an identified sparse dynamical system may be more useful than a highly accurate but non-interpretable surrogate.

Previous approaches combining dynamical system discovery with multi-fidelity data have often relied on recurrent neural networks to learn a mapping from low-fidelity observations to high-fidelity dynamics, so that prediction at inference time remains dependent on evaluating a low fidelity model \cite{conti_mf}. Other SINDy-based approaches use multi-fidelity Gaussian processes for noise smoothing and data fusion, thereby introducing an additional surrogate layer into the identification pipeline \cite{MENG2025113651}. More recently, weighted formulations have also been proposed to improve the efficiency of SINDy training \cite{colbrook}; however, these works do not investigate the role of weighting in the presence of multi-fidelity data.

\begin{figure}
    \centering
    \includegraphics[width=\textwidth]{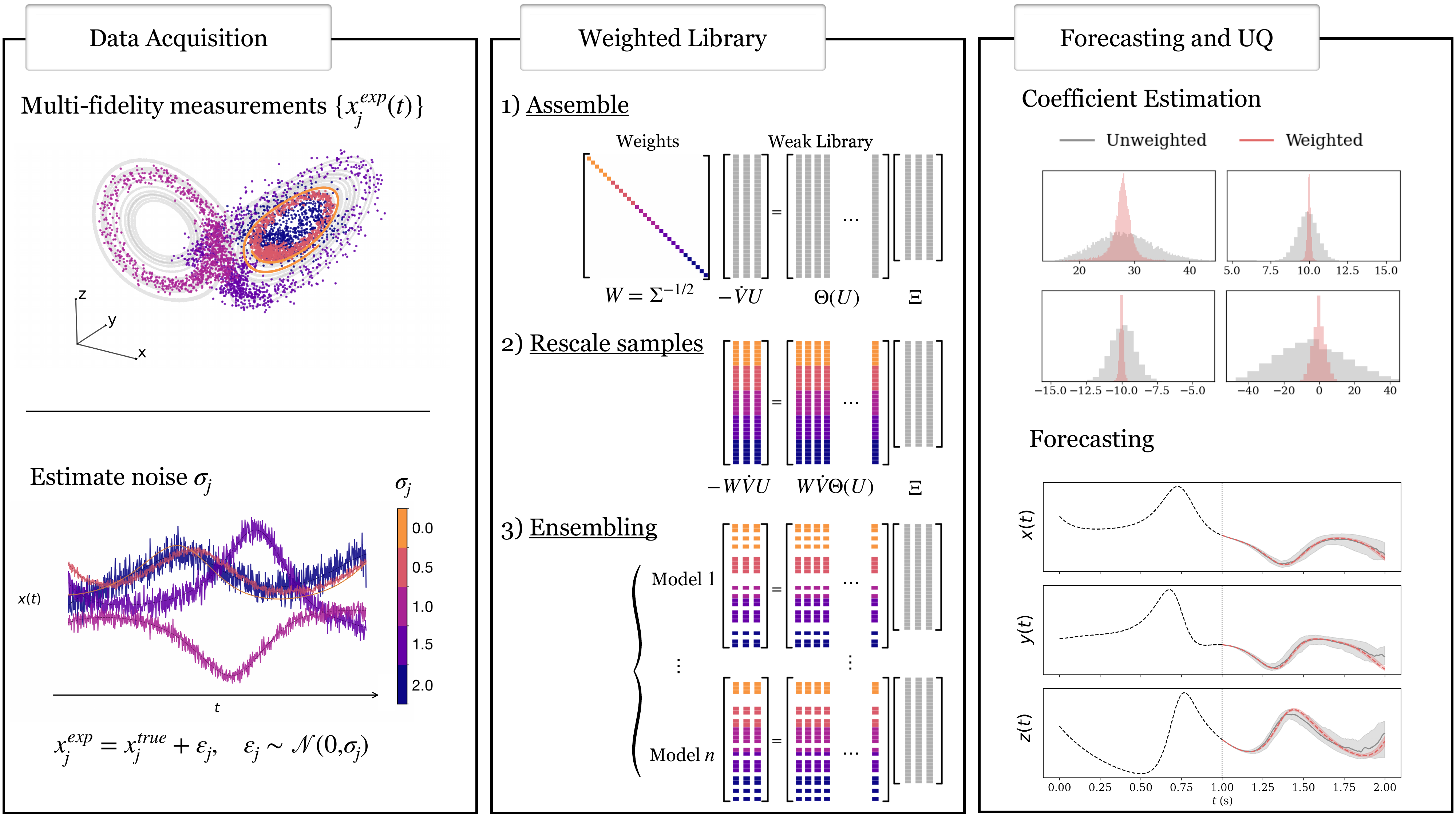}
    \caption{
    Schematic overview of MF-SINDy methodology for model discovery from multi-fidelity data.
    (1) \textbf{Data acquisition:} measurements are collected at multiple fidelity levels, and a heteroscedastic noise standard deviation is estimated for each sample or trajectory.
    (2) \textbf{Assembly and weighting:} the weak library matrix and right-hand side are assembled and reweighted using the covariance induced by heteroscedastic noise, yielding a GLS-weighted weak regression problem. The rows of the weighted weak system are repeatedly subsampled to form an ensemble of models. The matrix $\Sigma$ is obtained from the estimated $\sigma_j$ according to Equations \eqref{eq:traj_specific_covariance} and \eqref{eq:multi_traj_block_covariance}.
    (3) \textbf{Forecasting with uncertainty:} the ensemble of identified models is propagated to obtain forecasts and associated uncertainty bands.
} 
    \label{fig:method}
\end{figure}

In contrast, the proposed MF--SINDy framework incorporates fidelity information directly into the offline identification stage. A weak formulation mitigates noise amplification due to numerical differentiation, an ensemble procedure supports uncertainty quantification, and covariance-aware weighting provides a simple and principled treatment of heteroscedastic measurements within a single sparse discovery pipeline. The methodology is particularly effective when trajectories are collected with different sensing devices or under distinct experimental or simulation conditions, leading to trajectory-specific noise levels that should be combined in a principled way. It also applies naturally when the noise level varies within a trajectory over time, for example because of changing operating conditions or external disturbances. A schematic overview is given in Fig.~\ref{fig:method}.

The remainder of the paper is organized as follows. Section \ref{sec:background} reviews the relevant background, and Section \ref{sec:methodology} presents the proposed methodology. The results are then reported for three settings: the multi-trajectory case in Section \ref{subsec:res1}, the single-trajectory case with heterogeneous noise in Section \ref{subsec:res2}, and double-pendulum forecasting with multi-fidelity data in Section \ref{subsec:res3}. The paper concludes with a discussion in Section \ref{sec:discussion} and final remarks in Section \ref{sec:conclusion}.

\section{Background}\label{sec:background}

We begin by reviewing the main ideas underlying the sparse identification of nonlinear dynamics (\gls{sindy}), a regression framework for discovering governing equations from measurement data. We first present the standard formulation and then discuss extensions that improve robustness to noise, namely weak \gls{sindy} and ensemble methods. We also summarize the optimization procedures most relevant to this work, with particular attention to sequential thresholded least squares and generalized least squares.

\subsection{Sparse Identification of Dynamical Systems}

Consider a dynamical system $\mathbf{u}:[0,T]\to\mathbb{R}^d$ governed by the \gls{ode}
\begin{equation}\label{eq:ode}
    \dot{\mathbf{u}}(t)=\mathbf{F}(\mathbf{u}(t)),\qquad \mathbf{u}(0)=\mathbf{u}_0,
\end{equation}
or, more generally, a spatially dependent system $\mathbf{u}:\Omega\times[0,T]\to\mathbb{R}^d$ governed by the \gls{pde}
\begin{equation}\label{eq:pde}
    \mathbf{u}_t=\mathbf{N}\!\left(\mathbf{u},\mathbf{u}_x,\mathbf{u}_{xx},\ldots;\mathbf{x}\right),
\end{equation}
where the operator $\mathbf{F}$ in the ODE setting, or $\mathbf{N}$ in the PDE setting, is unknown.

Given measurements of the system state, the goal of \gls{sindy} is to identify an approximation of the right-hand side in \eqref{eq:ode} or \eqref{eq:pde}. This requires estimates of the time derivatives. Since these derivatives are usually not measured directly, they must be approximated from the data, for example by finite differences, smoothing procedures, or other numerical differentiation techniques~\cite{brunton2016}.

Let $\mathbf{u}_i=\mathbf{u}(t_i)$, for $i=1,\ldots,N$, denote the observed samples, and let $\dot{\mathbf{u}}_i$ denote the corresponding approximations of the time derivatives at time $t_i$. These quantities are collected in the matrices:
\begin{equation}
    \mathbf{U}=
    \begin{bmatrix}
        \mathbf{u}_1, \mathbf{u}_2, \ldots, \mathbf{u}_N
    \end{bmatrix}^\top
    \in\mathbb{R}^{N\times d},
    \qquad
    \mathbf{U}_t=
    \begin{bmatrix}
        \dot{\mathbf{u}}_1, \dot{\mathbf{u}}_2, \ldots, \dot{\mathbf{u}}_N
    \end{bmatrix}^\top
    \in\mathbb{R}^{N\times d}.
\end{equation}

The \gls{sindy} approach assumes that these derivatives can be approximated by a sparse linear combination of candidate functions evaluated on the observed data. To this end, a library of $D$ candidate terms is constructed,
\begin{equation}
    \boldsymbol{\Theta}(\mathbf{U})=
    \big[\theta_1(\mathbf{U}),\theta_2(\mathbf{U}),\ldots,\theta_D(\mathbf{U})\big]
    \in\mathbb{R}^{N\times D},
\end{equation}
including, for example, polynomials and trigonometric functions of $\mathbf{u}$ and, in the PDE case, spatial derivatives and their nonlinear combinations. The choice of the library is fundamental: it must be sufficiently rich to represent the dynamics of interest while remaining structured enough to allow sparse identification. A common strategy is to begin with low-order polynomials and then gradually enrich the library until accurate models are obtained.

SINDy seeks a sparse coefficient matrix $\boldsymbol{\Xi}$ such that the observed dynamics are approximated by a linear combination of candidate terms:
\begin{equation}
    \mathbf{U}_t \approx \boldsymbol{\Theta}(\mathbf{U})\,\boldsymbol{\Xi}.
\end{equation}
The coefficient matrix $\boldsymbol{\Xi} \in \mathbb{R}^{D\times d}$ contains the weights associated with the candidate terms, and the terms corresponding to nonzero coefficients define the active support.
Enforcing parsimony yields the sparse regression problem:
\begin{equation}\label{eq:sindy_regression}
    \widehat{\boldsymbol{\Xi}}
    =\arg\min_{{\boldsymbol{\Xi}}}
        \frac{1}{2}\big\|
            \mathbf{U}_t-\boldsymbol{\Theta}(\mathbf{U})\,{\boldsymbol{\Xi}}
        \big\|_F^2
        + R(\boldsymbol{\Xi}),
\end{equation}
where the regularizer $R(\widehat{\boldsymbol{\Xi}})$ is chosen to promote sparsity in the recovered coefficients.

\subsection{Optimization Routine}
\label{subsec:optimization}

The sparse regression problem in \eqref{eq:sindy_regression} is commonly solved with an iterative procedure based on \gls{stridge} \cite{rudy2017}, that alternates between two steps:
\begin{enumerate}
    \item a ridge-regularized least-squares update,
    \begin{equation}\label{eq:ls}
        \widehat{\boldsymbol{\Xi}}
        =
        \arg\min_{\boldsymbol{\Xi}}
        \big\|
            \mathbf{U}_t - \boldsymbol{\Theta}(\mathbf{U})\,\boldsymbol{\Xi}
        \big\|_F^2
        + \lambda_2 \|\boldsymbol{\Xi}\|_F^2,
    \end{equation}
    \item a hard-thresholding step, in which all coefficients satisfying \( |\Xi_{ij}| < \lambda \) are set to zero and kept fixed in subsequent iterations.
\end{enumerate}
These two steps are repeated until the support of the coefficient matrix no longer changes, that is, until the set of nonzero entries of \(\boldsymbol{\Xi}\) remains unchanged across successive iterations. The threshold parameter $\lambda$ controls the sparsity level of the identified model, whereas the ridge parameter $\lambda_2$ improves the conditioning of the least-squares problem, especially in the presence of strongly correlated library terms. The classical \gls{stls} method, introduced in the original \gls{sindy} formulation \cite{brunton2016}, is recovered in the special case $\lambda_2=0$, for which the first step reduces to an unregularized least-squares problem. The rigorous formulation of the scheme is presented in Algorithm \ref{alg:stls}.

Beyond these standard approaches, alternative optimization strategies have been proposed to improve robustness and incorporate prior information into the identification process. For example, Loiseau and Brunton \cite{loiseau2018} introduced a constrained \gls{sr3} formulation in which the original sparse regression problem is relaxed through an auxiliary variable and a penalty term that separates sparsity regularization from data fitting, further enabling symmetry constraints to be incorporated. This often improves conditioning and makes it easier to impose additional linear constraints on the coefficient matrix. Kaptanoglu et al. \cite{kaptanoglu2021} proposed a stability-promoting formulation for quadratically nonlinear systems based on the trapping theorem \cite{trapping}, yielding identified models that satisfy boundedness conditions and therefore avoid divergent long-term behavior. Such approaches are particularly useful when physical prior knowledge must be incorporated to ensure that the identified dynamics remain interpretable and consistent with known system behavior.

\begin{algorithm}[t]
\caption{STRidge sparse regression}
\label{alg:stls}
\begin{algorithmic}
\STATE \textbf{Input:}  derivatives $\mathbf{U}_t\in \mathbb{R}^{N \times d}$, library matrix $\boldsymbol{\Theta}(\mathbf{U})$, threshold $\lambda$, ridge parameter $\lambda_2$
\STATE \textbf{Output:} sparse coefficient matrix $\widehat{\boldsymbol{\Xi}}$
\FOR{$r=1$ \TO $d$}
    \STATE Compute an initial ridge estimate
    $\widehat{\boldsymbol{\xi}}_r
    =
    \arg\min_{\boldsymbol{\xi}}
    \|(\mathbf{U}_t)_{\cdot r}-\boldsymbol{\Theta}(\mathbf{U})\boldsymbol{\xi}\|_2^2
    +\lambda_2\|\boldsymbol{\xi}\|_2^2$
    \STATE Define the support $\mathcal{S}_r=\{j:|\widehat{\xi}_{r,j}|\ge \lambda\}$
    \REPEAT
        \STATE Store the previous support $\mathcal{S}_r^{\mathrm{old}}=\mathcal{S}_r$
        \STATE Refit $\widehat{\boldsymbol{\xi}}_r$ by solving the same ridge problem restricted to the active support $\mathcal{S}_r$
        \STATE Threshold coefficients with magnitude below $\lambda$ and update the support $\mathcal{S}_r=\{j:|\widehat{\xi}_{r,j}|\ge \lambda\}$
    \UNTIL{$\mathcal{S}_r=\mathcal{S}_r^{\mathrm{old}}$}
\ENDFOR
\STATE Assemble $\widehat{\boldsymbol{\Xi}}=[\widehat{\boldsymbol{\xi}}_1,\ldots,\widehat{\boldsymbol{\xi}}_d]$
\end{algorithmic}
\end{algorithm}
\subsection{Noisy Data: Weak Formulations}\label{sec:weak}
In realistic acquisition settings, one observes noisy measurements rather than exact states. More precisely, instead of the true state \(\mathbf{u}(t_i)\), one observes
\begin{equation}
    \hat{\mathbf{u}}_i = \mathbf{u}(t_i) + \boldsymbol{\varepsilon}_i,
    \qquad 
    \boldsymbol{\varepsilon}_i \sim \mathcal{N}(\mathbf{0},\sigma^2\mathbf{I}_d),
    \label{eq:noisy}
\end{equation}
where the residuals are assumed to be i.i.d.\ Gaussian with constant variance and \(\mathbf{I}_d \in \mathbb{R}^{d\times d}\) denotes the identity matrix. 
Noise creates a major difficulty for standard \gls{sindy}, because the method typically requires estimates of time derivatives from measured trajectories. Numerical differentiation is highly sensitive to perturbations in the data, and finite-difference approximations may strongly amplify noise, especially when the sampling interval \(\Delta t\) is small. Measurement noise also propagates to the identified coefficients and to the resulting model predictions, making uncertainty in both model selection and forecasting non-negligible. These limitations motivate weak formulations, which avoid explicit pointwise differentiation, as well as ensemble-based strategies for uncertainty quantification and model selection \cite{fasel2022}.

The central idea of weak \gls{sindy} is to replace the pointwise dynamical relation with a family of integral identities obtained by testing the governing equation against smooth functions \cite{schaeffer2017,messenger2021_1,messenger2021_2}.

Let $\{\varphi_k\}_{k=1}^K$ be a collection of sufficiently smooth test functions, each supported on a subinterval $D_k \subset [0,T]$. Multiplying the governing equation by $\varphi_k$ and integrating over $D_k$ gives
\begin{equation}
    \int_{D_k} \dot{\mathbf{u}}(t)\,\varphi_k(t)\,\mathrm{d}t
    =
    \int_{D_k} \mathbf{F}(\mathbf{u}(t))\,\varphi_k(t)\,\mathrm{d}t,
    \qquad k=1,\dots,K.
\end{equation}
Applying integration by parts to the left-hand side yields
\begin{equation}
    \int_{D_k} \dot{\mathbf{u}}(t)\,\varphi_k(t)\,\mathrm{d}t
    =
    \left[\mathbf{u}(t)\varphi_k(t)\right]_{\partial D_k}
    -
    \int_{D_k} \mathbf{u}(t)\,\dot{\varphi}_k(t)\,\mathrm{d}t.
\end{equation}
If the test functions are chosen so that they vanish at the endpoints of their supports, the boundary contribution disappears, and one obtains
\begin{equation}
    -\int_{D_k} \mathbf{u}(t)\,\dot{\varphi}_k(t)\,\mathrm{d}t
    =
    \int_{D_k} \mathbf{F}(\mathbf{u}(t))\,\varphi_k(t)\,\mathrm{d}t
    \approx
    \left( \int_{D_k} \boldsymbol{\Theta}(\mathbf{u}(t))\,\varphi_k(t)\,\mathrm{d}t \right)\boldsymbol{\Xi},
    \qquad k=1,\dots,K.
    \label{eq:weak_ode_continuous}
\end{equation}

This formulation is advantageous because derivatives are transferred from the noisy state \(\mathbf{u}\) to the smooth test functions \(\varphi_k\). As a result, the regression involves weighted integrals of the data rather than pointwise derivative estimates, which substantially improves robustness to measurement noise.

The choice of test functions is important in practice. They must be sufficiently smooth to justify integration by parts and are typically chosen to be compactly supported or to vanish at the boundaries of their support so that boundary terms disappear. Common choices include polynomial, spline, or bump functions localized on overlapping time windows \cite{messenger2021_1,messenger2021_2}. Localized test functions are particularly useful because they preserve some temporal resolution while still providing the smoothing effect induced by integration.

Assume now that the trajectory is sampled on an equispaced grid $
    0=t_1 <  \dots < t_N = T,
   \ t_{i+1}-t_i = \Delta t.
$
Let $\widehat{\mathbf{U}} \in \mathbb{R}^{N\times d}$ denote the matrix of noisy measurements. 
Approximating the integrals in \eqref{eq:weak_ode_continuous} by a quadrature rule, for example the trapezoidal rule, yields the discrete matrices
\begin{equation}
    V_{ki} = w_i\,\varphi_k(t_i),
    \qquad
    \dot{V}_{ki} = w_i\,\dot{\varphi}_k(t_i),
\end{equation}
where $w_i$ are the quadrature weights. One then defines
\begin{equation}
    \mathbf{b} := -\dot{\textbf{V}}\,\widehat{\mathbf{U}} \in \mathbb{R}^{K\times d},
    \qquad
    \textbf{G} := \textbf{V}\,\boldsymbol{\Theta}(\widehat{\mathbf{U}}) \in \mathbb{R}^{K\times D},
\end{equation}
so that the weak formulation leads to the regression problem
\begin{equation}
    \mathbf{b} \approx \textbf{G} \boldsymbol{\Xi}.
    \label{eq:weak_discrete}
\end{equation}
Hence, the coefficient matrix $\boldsymbol{\Xi}$ is estimated by solving the sparse optimization problem
\begin{equation}
    \min_{\boldsymbol{\Xi}} \;
    \frac{1}{2}\left\|\mathbf{b}-\textbf{G}\boldsymbol{\Xi}\right\|_2^2
    +  R(\boldsymbol{\Xi}).
\end{equation}
Therefore, the standard pointwise regression based on estimated derivatives is replaced by an integral regression problem built from weighted averages of the data. Analogous optimization strategies may be employed, but the quantities entering it are more stable in the presence of noise. For instance the coefficients $\boldsymbol{\Xi}$ can be obtained by using Algorithm \ref{alg:stls} using as arguments $\mathbf{b}$ and $\mathbf{G}$ instead of $\mathbf{U}_t$ and $\boldsymbol{\Theta}(\mathbf{U})$.

In the PDE setting introduced in \eqref{eq:pde}, the same idea extends to space-time test functions with local supports.
Multiplying by $\varphi_k$ and integrating over spatio-temporal domains $D_k$ gives:
\begin{equation}
    \int_{D_k} \mathbf{u}_t(\mathbf{x},t)\,\varphi_k(\mathbf{x},t)\,\mathrm{d}\mathbf{x}\,\mathrm{d}t
    =
    \int_{D_k} \mathbf{N}\!\left(\mathbf{u},\mathbf{u}_x,\mathbf{u}_{xx},\ldots;\mathbf{x}\right)\,\varphi_k(\mathbf{x},t)\,\mathrm{d}\mathbf{x}\,\mathrm{d}t, \qquad k=1,\ldots,K.
\end{equation}
Integration by parts in time shifts the temporal derivative from $\mathbf{u}$ onto $\varphi_k$. Likewise, spatial derivatives appearing in the candidate terms can be transferred onto the test functions by repeated integration by parts~\cite{schaeffer2017,messenger2021_2}. For instance, if a diffusion term $\mathbf{u}_{xx}$ is present, integration by parts transfers the second spatial derivative from $\mathbf{u}$ onto the test:
\begin{equation}
    \int_{D_k} \mathbf{u}_{xx}(\mathbf{x},t)\,\varphi_k(\mathbf{x},t)\,\mathrm{d}\mathbf{x}\,\mathrm{d}t
    =
    \int_{D_k} \mathbf{u}(\mathbf{x},t)\,\partial_{xx}\varphi_k(\mathbf{x},t)\,\mathrm{d}\mathbf{x}\,\mathrm{d}t, \qquad k=1,\ldots,K, 
\end{equation}
up to boundary contributions that vanish under appropriate assumptions. More generally, all derivatives appearing in the PDE library can be shifted from the measured field onto the smooth test functions, thereby avoiding direct numerical differentiation of noisy data. 

After discretization in space and time, the PDE weak formulation leads once more to a matrix regression of the form \eqref{eq:weak_discrete}, 
where $\mathbf{b}$ collects the integrated temporal contributions and $\textbf{G}$ contains the integrated candidate-library terms evaluated against the space-time test functions. The difference is that now $\mathbf{V}$ includes additional terms coming from integrating the spatial derivatives by parts. Thus, the same sparse regression machinery can be applied in both ODE and PDE settings, with the main difference lying in the construction of the weak feature matrix.

In summary, weak formulations replace explicit derivative estimation by integral identities involving smooth test functions. This substantially reduces the noise amplification associated with numerical differentiation and makes sparse system identification more robust under realistic measurement conditions~\cite{messenger2021_1,messenger2021_2}. At the same time, the remaining variability induced by noise, finite sampling, and the choice of test functions motivates the use of complementary uncertainty quantification and model-selection strategies.

\subsection{Ensemble SINDy}
Ensemble \gls{sindy} (E-SINDy)~\cite{fasel2022} extends the basic framework by repeatedly fitting
\gls{sindy} models on resampled subsets of the data, producing an ensemble of coefficient
matrices \(
    \big\{\boldsymbol{\Xi}^{(k)}\big\}_{k=1}^{N_e}.
\)
This ensemble representation allows us to assess the reliability of individual terms and to quantify uncertainty in the identified coefficients and in downstream forecasts. 
Writing \(\boldsymbol{\Xi} = [\boldsymbol{\xi}_1,\ldots,\boldsymbol{\xi}_d]\), each
\(\boldsymbol{\xi}_i\) is treated as a random variable, and the ensemble provides
an empirical approximation to its distribution.  

From the ensemble, one may compute sample covariance matrices \(\operatorname{Cov}(\boldsymbol{\xi}_i)\) for each state component. The pointwise predictive uncertainty of
\(\mathbf{u}_t = \boldsymbol{\Theta}(\mathbf{u})\,\boldsymbol{\Xi}\)
at a given input \(\mathbf{u}\) can then be approximated by
\begin{equation}
    \operatorname{Tr}\bigl(
        \operatorname{Var}_{\boldsymbol{\Xi}}[\mathbf{u}_t \mid \mathbf{u}]
    \bigr)
    \;\approx\;
    \sum_{i=1}^d
        \boldsymbol{\Theta}(\mathbf{u})\,
        \operatorname{Cov}(\boldsymbol{\xi}_i)\,
        \boldsymbol{\Theta}(\mathbf{u})^\top.
\end{equation}
Importantly, E-SINDy is compatible with most SINDy variants, so the same ensemble machinery can be combined with weak formulations and weighted or generalized least squares, as considered in this work. An example of E-SINDy combined with \gls{stridge} is presented in Algorithm \ref{alg:ensemble_sindy}.

\begin{algorithm}[t]
\caption{Ensemble SINDy}
\label{alg:ensemble_sindy}
\begin{algorithmic}
\STATE \textbf{Input:} number of samples $N$, derivatives $\mathbf{U}_t$, library matrix $\boldsymbol{\Theta}(\mathbf{U})$, threshold $\lambda$, ridge parameter $\lambda_2$, number of ensemble members $N_e$, sampling fraction $\rho$
\STATE \textbf{Output:} ensemble of coefficient matrices $\{\widehat{\boldsymbol{\Xi}}^{(m)}\}_{m=1}^{N_e}$
\FOR{$m=1$ \TO $N_e$}
    \STATE Draw an index set $\mathcal{I}^{(m)}\subset\{1,\dots,N\}$ with sampling fraction $\rho$
    \STATE Set $\mathbf{U}_t^{(m)}=(\mathbf{U}_t)_{\mathcal{I}^{(m)},:}$ and $\boldsymbol{\Theta}(\mathbf{U})^{(m)}=\boldsymbol{\Theta}(\mathbf{U})_{\mathcal{I}^{(m)},:}$
    \STATE Compute $\widehat{\boldsymbol{\Xi}}^{(m)}$ by applying Algorithm~\ref{alg:stls}: \[ \widehat{\boldsymbol{\Xi}}^{(m)} = \text{Algoritm}\ref{alg:stls}(\mathbf{U}_t^{(m)},\boldsymbol{\Theta}(\mathbf{U})^{(m)},\lambda,\lambda_2)\]
\ENDFOR
\STATE Collect the ensemble $\{\widehat{\boldsymbol{\Xi}}^{(m)}\}_{m=1}^{N_e}$
\end{algorithmic}
\end{algorithm}

\subsection{Heteroscedastic Noise and Weighted Least Squares}
\label{subsec:wls_gls_background}
In both standard and weak \gls{sindy}, the identification step can be written, for each state component \(r=1,\ldots,d\), as a linear regression problem of the form
\begin{equation}
    \mathbf{y}_r = \mathbf{A}\boldsymbol{\xi}_r + \boldsymbol{\varepsilon}_r,
    \label{eq:generic_regression}
\end{equation}
where \(\mathbf{y}_r\in\mathbb{R}^n\) denotes the response vector, \(\mathbf{A}\in\mathbb{R}^{n\times D}\) is the feature matrix, \(\boldsymbol{\xi}_r\in\mathbb{R}^D\) is the coefficient vector to be estimated, and \(\boldsymbol{\varepsilon}_r\in\mathbb{R}^n\) is the regression residual. In the standard \gls{sindy} setting, \(\mathbf{y}_r\) corresponds to the estimated time derivative of the \(r\)th state component and \(\mathbf{A}\) is the library matrix \(\boldsymbol{\Theta}(\mathbf{U})\). In weak \gls{sindy}, \(\mathbf{y}_r\) is replaced by the corresponding weak response and \(\mathbf{A}\) by the weak feature matrix. Thus, the discussion below applies to both formulations.

In the simplest setting, the residual is assumed to be centered, independent across observations, and homoscedastic:
\begin{equation}
    \boldsymbol{\varepsilon}_r \sim \mathcal{N}(\mathbf{0},\sigma^2\mathbf{I}).
    \label{eq:homoscedastic_residual}
\end{equation}
Under this Gaussian model, ordinary least squares coincides with the maximum-likelihood estimator and, under the usual Gauss–Markov assumptions, is also the \gls{blue} \cite{charnes_equivalence_1976,shaffer_gauss_markov_1991}, i.e., the linear unbiased estimator with minimum variance. Sparse methods used in \gls{sindy}, such as \gls{stls} and \gls{stridge}, build on this least-squares formulation by combining it with thresholding or regularization to promote sparse coefficient vectors.

In many applications, however, the residuals may exhibit non-constant variance and may also be correlated. This leads to the more general covariance structure
\begin{equation}
    \mathrm{Cov}(\boldsymbol{\varepsilon}_r)=\boldsymbol{\Sigma},
    \qquad
    \boldsymbol{\Sigma}\neq \sigma^2 \mathbf{I}.
    \label{eq:Sigma_general}
\end{equation}
In this case, ordinary least squares remains unbiased under standard assumptions, but is no longer efficient. A natural alternative is generalized least squares, which explicitly accounts for the covariance of the residuals. To account for this covariance structure, one introduces a matrix $\mathbf{W}$ such that
\begin{equation}
    \mathbf{W}^\top \mathbf{W}=\boldsymbol{\Sigma}^{-1},
\end{equation}
for instance by taking $\mathbf{W}$ as a Cholesky factor of $\boldsymbol{\Sigma}^{-1}$. Premultiplying the regression equation by $\mathbf{W}$ yields a transformed problem with uncorrelated residuals of unit covariance. The coefficient vector is then estimated by solving
\begin{equation}
    \widehat{\boldsymbol{\xi}}_r
    =
    \arg\min_{\boldsymbol{\xi}\in\mathbb{R}^{D}}
    \left\|
        \mathbf{W}\bigl(\mathbf{y}_r-\mathbf{A}\boldsymbol{\xi}\bigr)
    \right\|_2^2.
    \label{eq:gls_component}
\end{equation}
Equivalently, generalized least squares can be interpreted as ordinary least squares applied to a whitened system. When the covariance matrix is correctly specified, it yields the \gls{blue} \cite{aitken1936}. 
In practice, the covariance matrix is rarely known a priori and must be estimated from the data. This leads to feasible generalized least squares, in which $\boldsymbol{\Sigma}$ is replaced by an estimate $\widehat{\boldsymbol{\Sigma}}$ \cite{greene2002econometricanalysis}. Although this introduces an additional estimation step, substantial efficiency gains may still be obtained when the estimated covariance captures the dominant heteroscedasticity or correlation structure.

In the following, we exploit this structure by introducing a generalized least-squares formulation for multi-fidelity samples and combining it with weak \gls{sindy}. In this setting, the regression response is obtained after integration against test functions, so its covariance reflects not only the underlying measurement noise but also the quadrature weights and the test functions themselves. The goal is to obtain a more statistically efficient identification procedure.
\section{Methodology}
\label{sec:methodology}

This section explains how to combine the ingredients introduced so far to develop a unified framework for sparse identification from heterogeneously noisy data. The goal is to identify nonlinear dynamical systems from measurements whose reliability may vary across trajectories, sensors, or time. To achieve this, we combine weak \gls{sindy}, which avoids explicit differentiation of noisy data, ensemble \gls{sindy}, which supports uncertainty quantification, and a covariance-aware weighting strategy accounting for heterogeneous noise levels. In particular, MF-SINDy proceeds in three stages, as illustrated in Fig.~\ref{fig:method}:
\begin{enumerate}
    \item \textbf{Data acquisition and fidelity annotation.}
    Measurements are collected across one or more fidelity levels and, when available, across multiple independent trajectories. Each sample, or local time interval, is associated with an estimated noise standard deviation, yielding a heteroscedastic variance profile that encodes measurement reliability.
    
    \item \textbf{Weak assembly and covariance-aware weighting.}
    A weak regression system is constructed by integrating the governing equations against localized test functions, thereby avoiding direct numerical differentiation of noisy measurements. The residual covariance induced by heteroscedastic noise is modeled explicitly in weak space and used to define a whitening operator, yielding a weighted weak regression problem.
    
    \item \textbf{Ensemble discovery and forecasting with uncertainty.}
    The whitened weak regression system is repeatedly subsampled to construct an ensemble of sparse models. The resulting empirical distribution over the coefficients is then propagated through the identified dynamics to obtain forecasts and associated uncertainty bands.
\end{enumerate}

Among these ingredients, the novelty lies in the weighting step and the synthesis of the various methodologies. Once the weak regression system has been assembled, the key question is how observations with different noise levels should contribute to regression. Our proposed approach addresses the question by examining the weak residual, deriving its covariance structure from the measurement noise, and using the resulting covariance matrix to whiten the regression system. This leads to a generalized least-squares formulation in weak space, after which standard sparse regression can be applied. 
Accordingly, the main task of this section is to construct the covariance matrix of the weak residual. We first consider the most general single-trajectory setting, in which the noise level may vary over time. We then extend the construction to multiple independent trajectories, where trajectory-specific covariance blocks arise naturally. Finally, we discuss the corresponding PDE extension and the practical estimation of the variance profile. 
Algorithm~\ref{alg:mfsindy_overview} rigorously summarizes the complete MF--SINDy procedure.

\subsection{Covariance-aware weak regression}
\label{subsec:covariance_rescaling}

Once the weak formulation has been assembled, the identification problem takes the form of a regression
\begin{equation}
    \mathbf{b}\approx \mathbf{G}\boldsymbol{\Xi},
\end{equation}
where \(\mathbf{b}\) and \(\mathbf{G}\) are built from integrated quantities. In the homoscedastic setting, this regression can be treated with the same least-squares-based sparse solvers used in standard \gls{sindy}. In the present setting, however, the measurements are affected by heterogeneous noise levels, and this heterogeneity propagates into the weak system. As a result, different weak constraints might have different reliability, and the corresponding regression error is generally neither homoscedastic nor uncorrelated.

The sparse identification step is performed on the weak system. Therefore, under a generalized least-squares perspective, the appropriate weighting must be derived from the covariance of the weak regression residual
    $\mathbf{R}(\boldsymbol{\Xi})=\mathbf{b}-\mathbf{G}\boldsymbol{\Xi}$. 
Once a covariance model 
    $\mathrm{Cov}\!\bigl(\mathbf{R}(\boldsymbol{\Xi}^\star)\bigr)\approx \boldsymbol{\Sigma}$ 
has been specified, the regression can be whitened by introducing a matrix \(\mathbf{W}\) such that
\begin{equation}\label{wMatrix}
    \mathbf{W}^\top\mathbf{W}=\boldsymbol{\Sigma}^{-1}.
\end{equation}
Applying \(\mathbf{W}\) to both sides of the weak system yields
\begin{equation}
    \mathbf{W}\mathbf{b}\approx \mathbf{W}\mathbf{G}\boldsymbol{\Xi},
\end{equation}
which approximately has identity covariance. The sparse regression step is then performed on this whitened system.

Accordingly, the central task is to characterize the covariance structure of the weak residual. The next subsections derive this covariance first for a single trajectory with time-varying noise, then for multiple independent trajectories, and finally for the PDE setting.

\subsection{Weak residual covariance for a single trajectory}
\label{subsec:single_traj_covariance}
We begin with a single trajectory observed on a time grid \(t_i=i\Delta t\), \(i=1,\ldots,N\).
The noisy measurements are assumed to satisfy
\begin{equation}
    \widehat{\mathbf{u}}(t_i)=\mathbf{u}(t_i)+\boldsymbol{\varepsilon}_i,
    \qquad
    \mathbb{E}[\boldsymbol{\varepsilon}_i]=\mathbf{0},
    \qquad
    \mathrm{Cov}(\boldsymbol{\varepsilon}_i)=\sigma_i^2\mathbf{I}_d,
    \label{eq:single_traj_noise}
\end{equation}
where the measurement perturbations are assumed to be independent across time. 
As in Section~\ref{sec:weak}, let \(\{\varphi_j\}_{j=1}^K\) be a family of localized test functions, and let \(\mathbf{V},\dot{\textbf{V}}\in\mathbb{R}^{K\times N}\) denote the associated quadrature matrices. Collecting the noisy samples in the matrix:
\begin{equation}
    \widehat{\mathbf{U}}
    =
    \begin{bmatrix}
    \widehat{\mathbf{u}}(t_1)^\top\\
    \vdots\\
    \widehat{\mathbf{u}}(t_N)^\top
    \end{bmatrix}
    \in\mathbb{R}^{N\times d},
\end{equation}
we assemble the weak-response and weak-feature matrices as
\begin{equation}
    \mathbf{b}:=-\dot{\textbf{V}}\widehat{\mathbf{U}}\in\mathbb{R}^{K\times d},
    \qquad
    \mathbf{G}:=\mathbf{V}\,\boldsymbol{\Theta}\!\big(\widehat{\mathbf{U}}\big)\in\mathbb{R}^{K\times D}.
    \label{eq:single_traj_weak_matrices}
\end{equation}
To characterize the covariance of this residual, write
\begin{equation}
    \widehat{\mathbf{U}}=\mathbf{U}+\mathbf{E},
\end{equation}
where \(\mathbf{U}\) denotes the noiseless trajectory and \(\mathbf{E}\) the measurement perturbation. As discussed in Appendix ~\ref{app:covariance_approx}, we adopt the leading-order approximation that the dominant stochastic contribution enters through the weak-response term \(\mathbf{b}\), whereas the perturbation induced in the weak feature matrix \(\mathbf{G}\) is smaller and is neglected at first order. Under this approximation,
\begin{equation}
    \mathbf{R}(\boldsymbol{\Xi}^\star)\approx -\dot{\textbf{V}}\mathbf{E}.
    \label{eq:single_traj_residual_approx}
\end{equation}
That is, to leading order, the weak residual is obtained by filtering the measurement perturbation through the derivative test operator \(\dot{\textbf{V}}\).
The covariance in weak space is then obtained by propagating the measurement covariance through this operator. If the measurement variances are collected in the diagonal matrix:
\begin{equation}
    \mathbf{D}_\sigma:=\mathrm{diag}(\sigma_1^2,\ldots,\sigma_N^2),
    \label{eq:single_traj_Dsigma}
\end{equation}
then, for each state component \(r\),
\begin{equation}
    \mathrm{Cov}\!\big(\mathbf{R}_{\cdot r}(\boldsymbol{\Xi}^\star)\big)
    \approx
    \boldsymbol{\Sigma}
    :=
    \dot{\textbf{V}}\mathbf{D}_\sigma(\dot{\textbf{V}})^\top, \qquad r=1,\ldots,d.
    \label{eq:single_traj_covariance}
\end{equation}
Even if the measurement perturbations are independent across time, the induced weak residuals are generally correlated because the corresponding weak equations involve overlapping test functions.

Once \(\boldsymbol{\Sigma}\) is known, let \(\mathbf{W}\) be a whitening matrix satisfying \eqref{wMatrix}, for example a Cholesky factor of \(\boldsymbol{\Sigma}^{-1}\). The covariance-aware sparse identification problem is then written in whitened form as
\begin{equation}
    \min_{\boldsymbol{\Xi}}
    \frac{1}{2}\bigl\|
        \mathbf{W}\mathbf{b}-\mathbf{W}\mathbf{G}\boldsymbol{\Xi}
    \bigr\|_F^2
    + R(\boldsymbol{\Xi}),
    \label{eq:single_traj_weighted_sparse}
\end{equation}
where \(R(\boldsymbol{\Xi})\) denotes the sparsity-promoting regularization or constraint. In practice, \eqref{eq:single_traj_weighted_sparse} may be solved with the same sparse optimization routines used in standard \gls{sindy}, after replacing the original weak system \((\mathbf{b},\mathbf{G})\) with the whitened system \((\mathbf{W}\mathbf{b},\mathbf{W}\mathbf{G})\). For example, under \gls{stls}, each least-squares refitting step is performed on the whitened system, and the thresholding step is then applied exactly as in the standard algorithm. The same substitution can be used in other \gls{sindy} solvers, such as \gls{stridge} or \gls{sr3}.

At the least-squares level, the whitening corresponds to a generalized least-squares weighting and therefore inherits its usual efficiency interpretation when the covariance model is correctly specified. Once sparsity-promoting regularization or thresholding is introduced, however, the \gls{blue} result no longer applies strictly, since the resulting estimator is no longer purely linear and unbiased. This inductive bias is introduced deliberately in order to recover parsimonious and interpretable governing equations. Nevertheless, accounting for the residual covariance before sparsification is still expected to improve the statistical quality of the recovered coefficients. 

\subsection{Extension to multiple trajectories}
\label{subsec:multi_traj_covariance}

We now extend the previous construction to multiple independent trajectories. Suppose that \(N_{\mathrm{traj}}\) trajectories are observed on a common time grid, and let \(\widehat{\mathbf{U}}^{(k)}\), \(\mathbf{b}^{(k)}\), and \(\mathbf{G}^{(k)}\) denote the weak matrices associated with trajectory \(k\), constructed exactly as in the single-trajectory setting.
For each trajectory, the leading-order weak residual is approximated by
\begin{equation}
    \mathbf{R}^{(k)}(\boldsymbol{\Xi}^\star)\approx -\dot{\textbf{V}}\mathbf{E}^{(k)},
\end{equation}
and therefore its covariance is
\begin{equation}
    \mathrm{Cov}\!\big(\mathbf{R}^{(k)}_{\cdot r}(\boldsymbol{\Xi}^\star)\big)
    \approx
    \boldsymbol{\Sigma}^{(k)}
    :=
    \dot{\textbf{V}}\mathbf{D}_{\sigma^{(k)}}(\dot{\textbf{V}})^\top,
    \label{eq:traj_specific_covariance}
\end{equation}
where \(\mathbf{D}_{\sigma^{(k)}}\) is the diagonal matrix collecting the measurement variances for trajectory \(k\).If the noise level is uniform along the \(k\)th trajectory, then \(\mathbf{D}_{\sigma^{(k)}}=\sigma_k^2\mathbf{I}\), and \eqref{eq:traj_specific_covariance} reduces to
\begin{equation}
    \boldsymbol{\Sigma}^{(k)}=\sigma_k^2\boldsymbol{\Sigma}_0,
    \qquad
    \boldsymbol{\Sigma}_0=\dot{\textbf{V}}(\dot{\textbf{V}})^\top.
\end{equation}
To combine all trajectories into a single identification problem, we stack the weak systems:
\begin{equation}
\mathbf{b}=
\begin{bmatrix}
\mathbf{b}^{(1)}\\
\vdots\\
\mathbf{b}^{(N_{\mathrm{traj}})}
\end{bmatrix},
\qquad
\mathbf{G}=
\begin{bmatrix}
\mathbf{G}^{(1)}\\
\vdots\\
\mathbf{G}^{(N_{\mathrm{traj}})}
\end{bmatrix},
\qquad
\mathbf{R}(\boldsymbol{\Xi})=\mathbf{b}-\mathbf{G}\boldsymbol{\Xi}.
\label{eq:stacked_multi_trajectory_system}
\end{equation}
Because the trajectories are independent, the stacked residual covariance is block diagonal. For each state component \(r\),
\begin{equation}
    \mathrm{Cov}\!\big(\mathbf{R}_{\cdot r}(\boldsymbol{\Xi}^\star)\big)
    \approx
    \boldsymbol{\Sigma}
    =
    \begin{bmatrix}
    \boldsymbol{\Sigma}^{(1)} & 0 & \cdots & 0\\
    0 & \boldsymbol{\Sigma}^{(2)} & \cdots & 0\\
    \vdots & \vdots & \ddots & \vdots\\
    0 & 0 & \cdots & \boldsymbol{\Sigma}^{(N_{\mathrm{traj}})}
    \end{bmatrix}.
    \label{eq:multi_traj_block_covariance}
\end{equation}
Thus, the multi-trajectory problem has exactly the same generalized least-squares structure as the single-trajectory problem; the only difference is that the covariance matrix now carries one block per trajectory.

This form makes the role of measurement fidelity transparent. A trajectory with larger overall noise variance contributes a larger covariance block and is therefore assigned less weight in the regression. More generally, if the noise level varies both across trajectories and within each trajectory, then the weighting adapts to both sources of heterogeneity simultaneously.

Generalized least squares is again implemented by introducing a whitening matrix \(\mathbf{W}\) satisfying \eqref{wMatrix}. The weighted sparse identification step is then carried out on the whitened system \((\mathbf{W}\mathbf{b},\mathbf{W}\mathbf{G})\). In the special case where each trajectory has a constant variance \(\sigma_k^2\), the stacked covariance matrix is block diagonal with blocks \(\sigma_k^2\boldsymbol{\Sigma}_0\). A corresponding block-diagonal whitening matrix is
\begin{equation}
\mathbf{W}=
\begin{bmatrix}
\sigma_1^{-1}\boldsymbol{\Sigma}_0^{-1/2} & 0 & \cdots & 0\\
0 & \sigma_2^{-1}\boldsymbol{\Sigma}_0^{-1/2} & \cdots & 0\\
\vdots & \vdots & \ddots & \vdots\\
0 & 0 & \cdots & \sigma_{N_{\mathrm{traj}}}^{-1}\boldsymbol{\Sigma}_0^{-1/2}
\end{bmatrix},
\label{eq:multi_traj_whitening_homo}
\end{equation}
where \(\boldsymbol{\Sigma}_0^{-1/2}\) denotes any matrix such that
\(
(\boldsymbol{\Sigma}_0^{-1/2})^\top \boldsymbol{\Sigma}_0^{-1/2}=\boldsymbol{\Sigma}_0^{-1}.
\)
Thus, whitening acts independently across trajectories, while preserving the within-trajectory correlation structure induced by the weak test functions.

\subsection{PDE extension}
\label{subsec:pde_covariance_extension}

The same covariance construction extends to PDE identification, with one important clarification. As discussed in Section~\ref{sec:weak}, after integration by parts in time and space, the weak PDE system still takes the form \eqref{eq:weak_discrete}, where the vector \(\mathbf{b}\) collects the integrated temporal contributions, while the matrix \(\mathbf{G}\) contains the candidate-library terms evaluated against space-time test functions, including the additional contributions generated by integration by parts of spatial derivatives.

As in the ODE case, we approximate the weak residual covariance by retaining only the leading noise contribution entering through the left-hand side term \(\mathbf{b}\), and neglecting at first order the perturbation induced in \(\mathbf{G}\). Let \(\mathbf{u}:\Omega\times[0,T]\to\mathbb{R}\) be observed on a space-time grid \(\{(\mathbf{x}_m,t_n)\}\), with noisy measurements
\begin{equation}
    \widehat{\mathbf{u}}_{m,n}=\mathbf{u}(\mathbf{x}_m,t_n)+\boldsymbol{\varepsilon}_{m,n},
    \qquad
    \mathbb{E}[\boldsymbol{\varepsilon}_{m,n}]=0,
    \qquad
    \mathrm{Var}(\boldsymbol{\varepsilon}_{m,n})=\sigma_{m,n}^2.
    \label{eq:pde_noise_model}
\end{equation}
If \(\mathbf{D}_{\sigma}^{\mathrm{st}}\) denotes the diagonal covariance matrix of the residuals vectorized over the space-time grid, then the corresponding weak residual covariance is
\begin{equation}
    \boldsymbol{\Sigma}_{\mathrm{pde}}
    \approx
    \dot{\textbf{V}}\,\mathbf{D}_{\sigma}^{\mathrm{st}}\,(\dot{\textbf{V}})^\top.
    \label{eq:pde_weak_covariance}
\end{equation}
This is the PDE analogue of the ODE covariance model: the measurement variances are propagated into weak space through the operator associated with the temporal left-hand side, whereas the right hand side perturbations are neglected.

This approximation is more delicate than in the ODE case. Because the PDE library may contain spatial derivatives and nonlinear spatial terms, the perturbation induced in \(\mathbf{G}\) is not negligible under arbitrary choices of test-function support. Accordingly, \eqref{eq:pde_weak_covariance} should be interpreted as a leading-order approximation under an additional scale-separation assumption. In particular, if the spatial support of the test functions is sufficiently large relative to their temporal support, spatial averaging suppresses the noise propagated through the library evaluation more strongly than the derivative-driven perturbation retained in \(\mathbf{b}\). Under this regime, the generalized least-squares formulation remains appropriate in the PDE setting as well. The corresponding assumptions are discussed in Appendix~\ref{app:covariance_approx}.

\subsection{Remark on variance specification and estimation}
\label{subsec:variance_estimation}

In practice, the variance profile is not always known a priori and must be estimated from the data. A simple strategy is to compare the raw trajectory with a smoothed version. Let \(\tilde{\mathbf{u}}(t_i)\) denote a denoised trajectory obtained, for example, by local polynomial smoothing. One may then define the residuals 
\begin{equation}
    \mathbf{e}_i=\widehat{\mathbf{u}}(t_i)-\tilde{\mathbf{u}}(t_i).
    \label{eq:residuals_variance_est}
\end{equation}
The local noise level can then be estimated by smoothing the squared residual norms:
\begin{equation}
    \widehat{\sigma}_i^2 \approx \mathcal{S}\!\left(\|\mathbf{e}_i\|_2^2\right),
    \label{eq:variance_estimator}
\end{equation}
where \(\mathcal{S}\) denotes a suitable smoothing operator. This is a standard strategy in heteroscedastic regression \cite{ruppert}. If a Savitzky-Golay smoother is used \cite{Steinier1964SmoothingAD}, the resulting variance estimates depend on the filter parameters and should therefore be interpreted with care 
\cite{oxby2024confidenceintervalssavitzkygolayfilter}.

The same idea carries over to PDE settings. There, one replaces temporal smoothing by an appropriate space-time smoother and forms residuals between the measured and smoothed fields. These residuals may then be used to estimate spatially and temporally varying noise levels, which are propagated into weak space through the same covariance construction described above. Although the details depend on the discretization and on the chosen smoother, the overall logic remains the same: the variance model is estimated in measurement space and then mapped into weak space through the test-function construction.

\begin{algorithm}[t]
\caption{MF-SINDy}
\label{alg:mfsindy_overview}
\begin{algorithmic}
\STATE \textbf{Input:} multi-fidelity trajectories $\{\widehat{\mathbf U}^{(k)}\}$, candidate library $\boldsymbol{\Theta}$, test functions $\{\varphi_j\}_{j=1}^K$, variance profiles $\{\widehat{\sigma}^{(k)}\}$, threshold $\lambda$, ridge parameter $\lambda_2$, number of ensemble members $N_e$, sampling fraction $\rho$
\STATE \textbf{Output:} ensemble of coefficient matrices $\{\widehat{\mathbf \Xi}^{(m)}\}_{m=1}^{N_e}$

\FOR{each trajectory $k$}
    \STATE Assemble the weak system
    \[
    \mathbf b^{(k)}=-\mathbf V' \widehat{\mathbf U}^{(k)},
    \qquad
    \mathbf G^{(k)}=\mathbf V\,\boldsymbol{\Theta}(\widehat{\mathbf U}^{(k)})
    \]
    \STATE Construct the trajectory covariance
    \[
    \mathbf \Sigma^{(k)}=\mathbf V' \mathbf D_{\sigma^{(k)}} (\mathbf V')^\top
    \]
\ENDFOR

\STATE Stack the weak systems and form the block covariance matrix:
\[
\mathbf b=
\begin{bmatrix}
\mathbf b^{(1)}\\ \vdots\\ \mathbf b^{(N_{\mathrm{traj}})}
\end{bmatrix},
\qquad
\mathbf G=
\begin{bmatrix}
\mathbf G^{(1)}\\ \vdots\\ \mathbf G^{(N_{\mathrm{traj}})}
\end{bmatrix},
\qquad
\mathbf \Sigma=
\begin{bmatrix}
\mathbf \Sigma^{(1)} &        & 0\\
                      & \ddots &  \\
0                     &        & \mathbf \Sigma^{(N_{\mathrm{traj}})}
\end{bmatrix}
\]

\STATE Compute a whitening matrix $\mathbf W$ from a Cholesky factorization of $\mathbf \Sigma^{-1}$, i.e.
\[
\mathbf W^\top \mathbf W=\mathbf \Sigma^{-1}
\]

\STATE Form the whitened weak system
\[
\widetilde{\mathbf b}=\mathbf W\mathbf b,
\qquad
\widetilde{\mathbf G}=\mathbf W\mathbf G
\]

\STATE Compute the ensemble of coefficient matrices by applying Algorithm~\ref{alg:ensemble_sindy}:
\[
\{\widehat{\mathbf \Xi}^{(m)}\}_{m=1}^{N_e}
=
\text{Algorithm~\ref{alg:ensemble_sindy}}(\widetilde{\mathbf b},\widetilde{\mathbf G},\lambda,\lambda_2,N_e,\rho)
\]
\end{algorithmic}
\end{algorithm}
\section{Results}\label{sec:results}

We evaluate the proposed framework in three settings.
Part~I studies model identification from multiple trajectories observed at two fidelity levels. In this setting, noise is homoscedastic within each trajectory but differs across fidelity groups.  
Part~II considers identification from a single trajectory showing within-trajectory heteroscedastic noise.  
Part~III shifts the focus from coefficient recovery to prediction and studies short-horizon forecasting for a planar double-pendulum benchmark.

The identification benchmarks (Part I and Part II) include five systems: the Lorenz system, Burgers' equation, the Hopf oscillator, the linear pendulum, and isothermal compressible flow. For these cases, we evaluate coefficient accuracy and support recovery relative to the reference model. The double-pendulum benchmark instead evaluates predictive accuracy and uncertainty quality.

For the identification experiments, let \(\boldsymbol{\Xi}^\star \in \mathbb{R}^{m \times n}\) denote the ground-truth coefficient matrix and \(\widehat{\boldsymbol{\Xi}} \in \mathbb{R}^{m \times n}\) the identified coefficient matrix. We report the mean absolute coefficient error
\begin{equation}
L_1(\widehat{\boldsymbol{\Xi}}, \boldsymbol{\Xi}^\star)
=
\frac{1}{mn}\sum_{i=1}^{m}\sum_{j=1}^{n}
\left| \widehat{\boldsymbol{\Xi}}_{ij} - \boldsymbol{\Xi}^\star_{ij} \right|,
\end{equation}
and the normalized support mismatch
\begin{equation}
L_0(\widehat{\boldsymbol{\Xi}}, \boldsymbol{\Xi}^\star)
=
\frac{\#\left( \widehat{\mathcal S} \ominus \mathcal S^\star \right)}
{\#\left( \mathcal S^\star \right)},
\qquad
\widehat{\mathcal S} = \operatorname{supp}(\widehat{\boldsymbol{\Xi}}),
\quad
\mathcal S^\star = \operatorname{supp}(\boldsymbol{\Xi}^\star),
\end{equation}
where \(\#(\cdot)\) denotes set cardinality and \(\ominus\) the symmetric difference. Thus, \(L_0\) measures support disagreement relative to the number of active terms in the reference model.

\subsection{Part I: Multi-Trajectory Multi-Fidelity Identification}
\label{subsec:res1}
\begin{figure}[t]
    \centering
    \includegraphics[width=\textwidth]{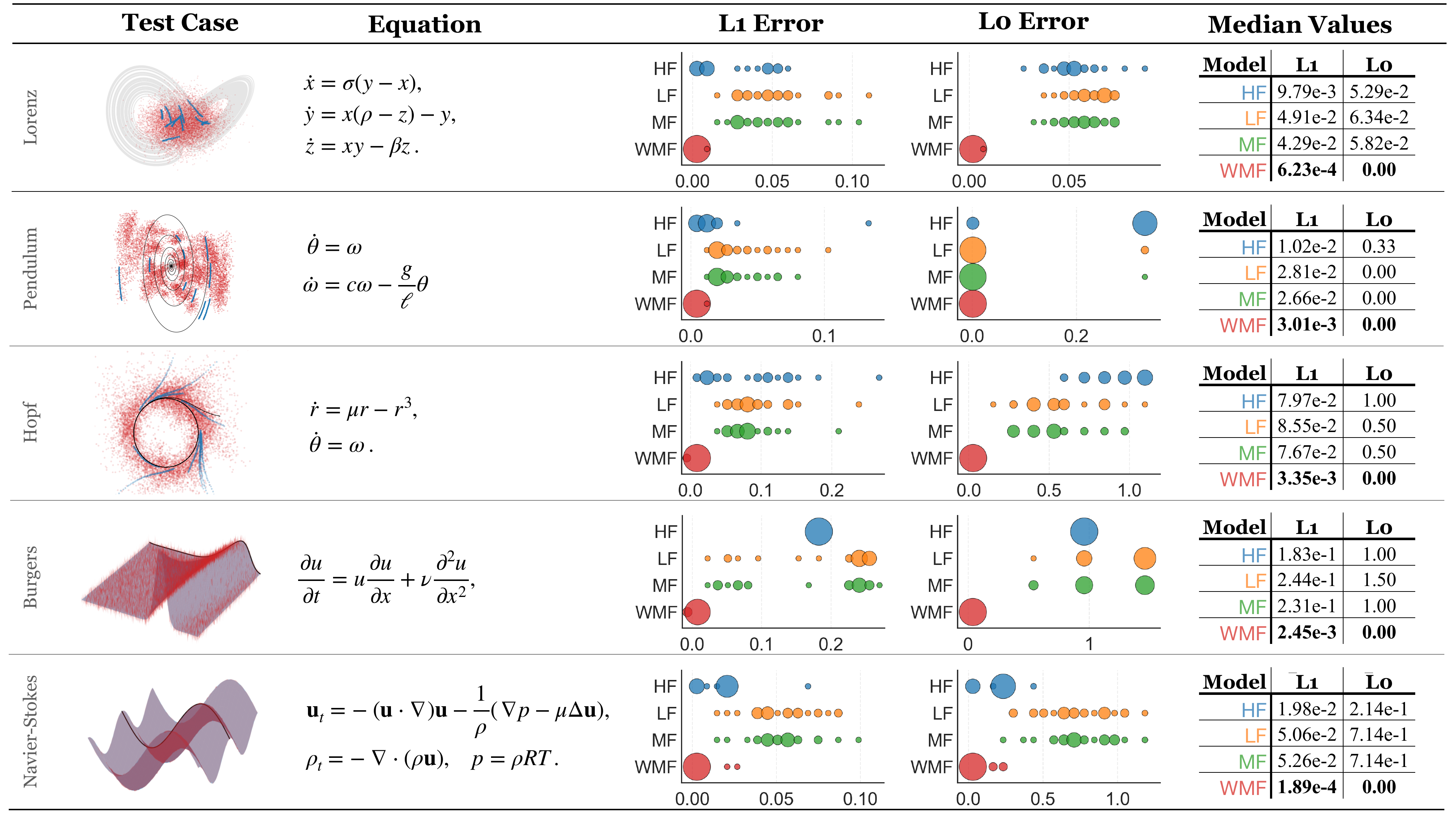}
    \caption{Part~I: multi-trajectory multi-fidelity identification across the five benchmark systems. Each row corresponds to one benchmark. The leftmost columns show the governing dynamics. The blue dots represent HF trajectories while red is for LF. The middle columns report the empirical distributions of the coefficient error \(L_1\) and support error \(L_0\), and the rightmost column lists the corresponding median values for \texttt{HF}, \texttt{LF}, \texttt{MF}, and \texttt{WMF}. Lower values are better in all quantitative panels.}
    \vspace{-0.5cm}
    \label{fig:part1-results}
\end{figure}

In Part~I, each ODE/PDE training set consists of 10 high-fidelity (HF) trajectories and 100 low-fidelity (LF) trajectories generated from distinct initial conditions under the same governing equations. Each trajectory is simulated over a time horizon of \(0.1\) seconds and sampled with time step \(\Delta t = 0.001\) seconds. All trajectories are produced numerically using a sufficiently accurate finite-difference scheme. The two fidelity levels differ only in the observation noise: HF trajectories are corrupted with additive zero-mean Gaussian noise at a \(1\%\) noise-to-signal ratio, whereas LF trajectories are corrupted with noise at a level 25 times larger. 

We compare four identification strategies, denoted \texttt{HF}, \texttt{LF}, \texttt{MF}, and \texttt{WMF}. The \texttt{HF} and \texttt{LF} strategies use only single-fidelity data, that is, either the HF or the LF trajectories alone, and identification is performed using Weak Ensemble SINDy. The \texttt{MF} strategy combines the HF and LF weak systems without covariance-aware weighting nor any way of prioritizing HF data. The \texttt{WMF} (Weighted Multi Fidelity) strategy instead applies the proposed covariance-aware framework and whitens each weak block using the rescaling described in Section~\ref{subsec:multi_traj_covariance}. All four strategies use the same candidate library, weak test functions, and sparse ensemble solver. The comparison therefore isolates the effect of how information from different fidelity levels is combined. 

For each strategy, the experiment is repeated 25 times by generating new HF and LF observations and re-running the identification procedure, yielding empirical distributions for the \(L_0\) and \(L_1\) errors. In all benchmarks, the library contains the true terms: for ODE systems, we use a polynomial library, while for PDEs we additionally include spatial derivatives of different orders. The main hyperparameters, i.e. the weak integration support and the thresholding parameter are selected by grid search.

Figure~\ref{fig:part1-results} presents the overall results. The LF and HF trajectories, together with the corresponding equations for each system, are shown in the left columns, while the approximated error distributions are reported on the right. The figure shows a consistent pattern across the five benchmarks. The \texttt{LF} strategy generally yields the broadest error distributions, indicating that a larger number of trajectories does not by itself compensate for lower data quality. The unweighted multi-fidelity strategy \texttt{MF} often closely mirrors the performance of \texttt{LF}, suggesting that, without covariance-aware scaling, the regression is dominated by the more numerous and noisier LF observations. By contrast, the covariance-aware strategy \texttt{WMF} achieves the lowest median errors in most cases and, more importantly, the most reliable support recovery.

Overall, these results show that the inclusion of multiple data sources is not automatically beneficial. Simply combining HF and LF information can lead to limited gains, or even to performance dominated by the lower-quality data. In contrast, covariance-aware scaling enables effective multi-fidelity fusion, improving both identification accuracy and the uncertainty of the recovered models.

\subsection{Part II: Single-Trajectory Identification with heteroscedastic Noise}
\label{subsec:res2}

\begin{figure}[t]
    \centering
    \includegraphics[width=\textwidth]{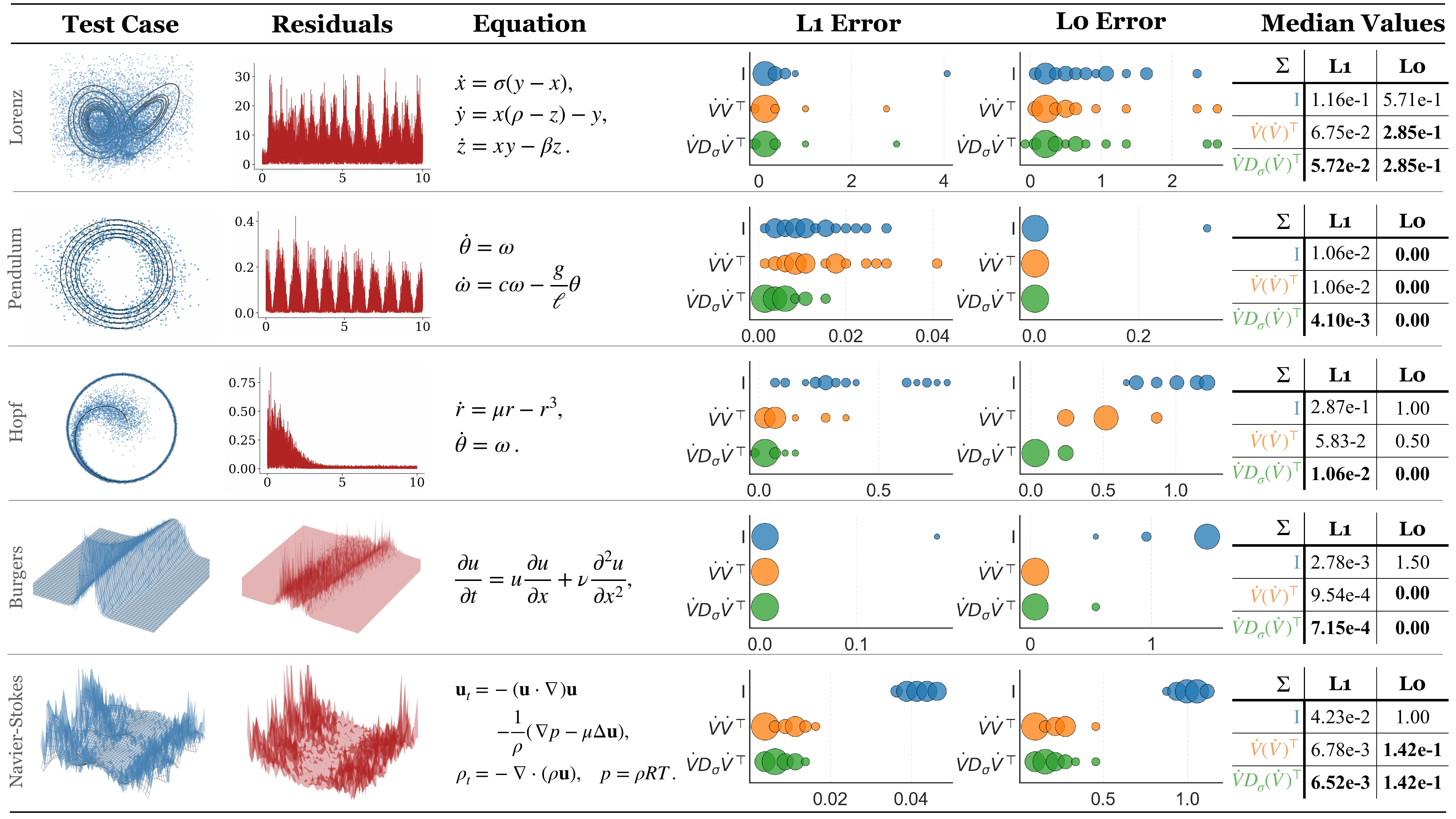}
    \caption{Part~II: single-trajectory identification with within-trajectory heteroscedastic noise across the five benchmark systems. The left columns illustrate representative trajectories and the profiles of the magnitude of the residuals in time, while the right columns report the distributions of \(L_1\) and \(L_0\) for three weak-space weighting choices: identity (no weighting), homogeneous-variance GLS induced by \(\dot{{V}}(\dot{{V}})^\top\) (accounting only for noise correlation), and heteroscedastic GLS induced by \(\dot{{V}} D_\sigma (\dot{{V}})^\top\), as described in sec \ref{subsec:single_traj_covariance}. Lower values are better in all quantitative panels.}
    \label{fig:part2-results}
    \vspace{-0.5cm}
\end{figure}

In Part~II, we consider the same benchmark systems introduced in Part~I, but now restrict the identification setting to a single observed trajectory of duration \(10\) seconds for each system. The data are generated using the same numerical schemes as in Part~I, and the sampling time is again set to \(0.001\) seconds. 
Training data are generated by corrupting the clean trajectory with heteroscedastic additive Gaussian noise:
\begin{equation}
\hat{\mathbf{u}}_i=\mathbf{u}(t_i)+\boldsymbol{\varepsilon}_i,
\qquad
\boldsymbol{\varepsilon}_i \sim \mathcal{N}\!\left(\mathbf{0},\sigma_i^2\right),
\end{equation}
where the noise variance varies over time and depends on the local state of the system. In this way, different portions of the trajectory are affected by different levels of uncertainty, producing a nonuniform reliability profile along the observation window.

The heteroscedastic noise profile is chosen separately for each benchmark so as to reflect a physically plausible state-dependent notion of noise intensity. The specific forms used in Part~II and their interpretations are summarized in Table~\ref{tab:part2_noise}. Representative trajectories and the associated residual profiles are shown in the first two columns of Figure~\ref{fig:part2-results}. These noise profiles should be interpreted as physically plausible heteroscedastic benchmark models rather than exact sensor models for the individual systems. Their role is to induce state-dependent observation reliability in a controlled and interpretable way, consistent with the broader framework of multiplicative or state-dependent noise in stochastic modeling \cite{gardiner2009stochastic}.

\begin{table}[t]
\centering
\caption{State-dependent heteroscedastic noise profiles used in Part~II.}
\label{tab:part2_noise}
\begin{tabular}{lll}
\hline
System & Noise standard deviation & Interpretation \\
\hline
Lorenz & $\sigma_i = 0.1\|\mathbf{u}(t_i)\|$ & proportional to state magnitude \\
Linear pendulum & $\sigma_i = 0.1\|\mathbf{u}(t_i)\|$ & proportional to state magnitude \\
Hopf oscillator & $\sigma_i = 0.1\,\bigl|\|\mathbf{u}(t_i)\|-1\bigr|$ & distance from limit cycle \\
Burgers' equation & $\sigma_i = 0.1\|u_x(x,t_i)\|$ & proportional to spatial gradient \\
Isothermal compressible flow & $\sigma_i = 0.1\|\mathbf{u}(\mathbf{x},t_i)\|$ & proportional to velocity magnitude \\
\hline
\end{tabular}
\end{table}

We compare three weak-space weighting strategies, all coupled with the same Weak Ensemble SINDy identification procedure: unweighted regression, unit-variance GLS, and heteroscedastic GLS. These strategies correspond to the generalized least-squares formulation introduced in Section~\ref{subsec:covariance_rescaling}, but differ in the choice of the covariance matrix \(\boldsymbol{\Sigma}\). The first uses the standard weak regression formulation without covariance correction, that is, \(\boldsymbol{\Sigma}\) is taken as the identity. The second uses the covariance structure \(\boldsymbol{\Sigma} = \dot{\textbf{V}}\dot{\textbf{V}}^\top\), thereby accounting for the correlation induced by overlapping weak test functions while assuming unit pointwise variance. The third uses the full heteroscedastic GLS construction of Equation~\eqref{eq:single_traj_covariance}, and therefore accounts for both weak-space correlation and the estimated nonuniform variance profile.

As in Part~I, the three weighting strategies are compared under the same identification pipeline so that the effect of covariance treatment can be isolated. The candidate library, weak test functions, sparse regression routine, and evaluation metrics are kept fixed across all cases. The hyperparameters, including the weak integration support and the thresholding parameter are selected by grid search. For each benchmark, the full identification procedure is repeated 25 times in order to obtain empirical distributions of the \(L_1\) and \(L_0\) errors, reported in Figure~\ref{fig:part2-results}. The corresponding distributions are shown in the fourth and fifth columns, while the median values are summarized in the last column.

The results show a clear separation between the unweighted baseline and the two covariance-aware alternatives. Across the five benchmark systems, unweighted regression is generally the least reliable option, leading to larger coefficient errors and less stable support recovery. Both unit-variance GLS and heteroscedastic GLS improve on this baseline, indicating that a substantial part of the gain already comes from accounting for the correlation structure induced by overlapping weak test functions. 
Across all benchmarks, unit-variance GLS already captures a large portion of the improvement over unweighted regression. This suggests that, even without explicitly modeling the variance field, correcting for weak-space correlation alone can substantially stabilize the regression problem. However, heteroscedastic GLS is generally the strongest performer.

Overall, the results of Part~II show that covariance-aware weighting is beneficial even in the single-trajectory setting. Accounting only for weak-space correlation already improves robustness relative to the unweighted baseline, but the best performance is obtained when the full heteroscedastic covariance structure is incorporated.
\begin{figure}[t]
    \centering
    \begin{minipage}[c]{0.38\textwidth}
        \centering
        \includegraphics[width=\textwidth]{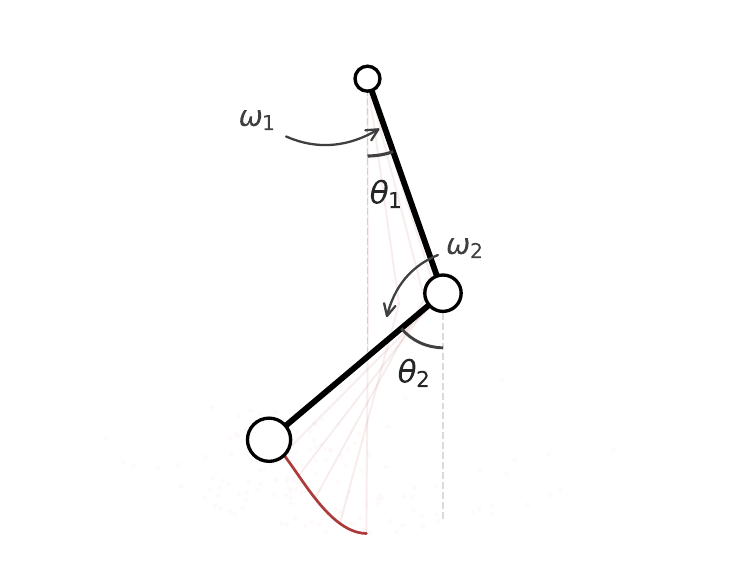}
    \end{minipage}
    \hfill
    \begin{minipage}[c]{0.48\textwidth}
        \centering
        \includegraphics[width=\textwidth]{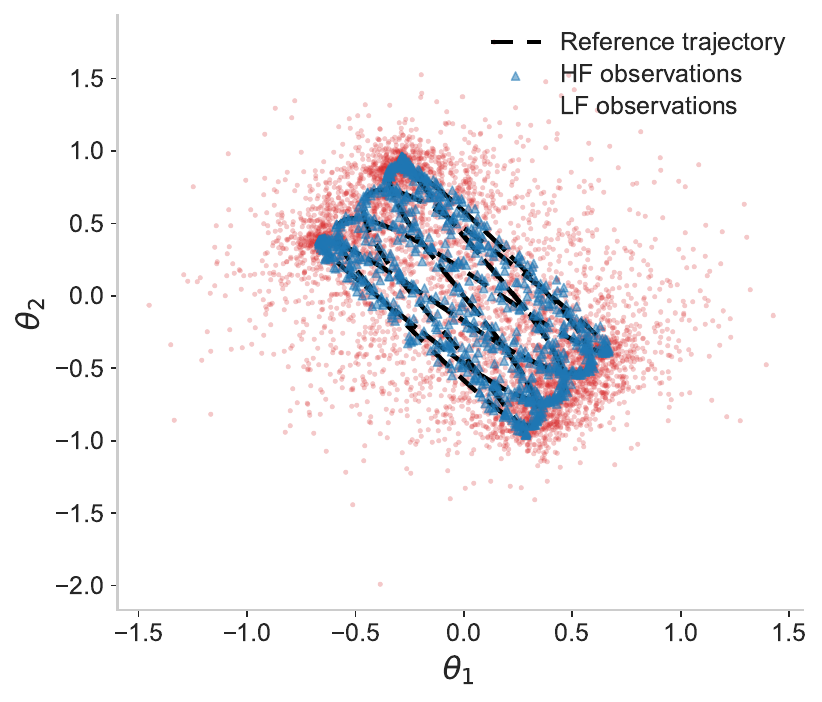}
    \end{minipage}
    \caption{Double-pendulum benchmark. Left: observation setup in Cartesian space. Right: configuration-space representation in the \((\theta_1,\theta_2)\) plane showing the reference trajectory together with high-fidelity and low-fidelity observations.}
    \vspace{-0.4cm}
    \label{fig:double-pendulum-setup}
\end{figure}
\begin{figure}[t]
    \centering
    \includegraphics[width=\textwidth]{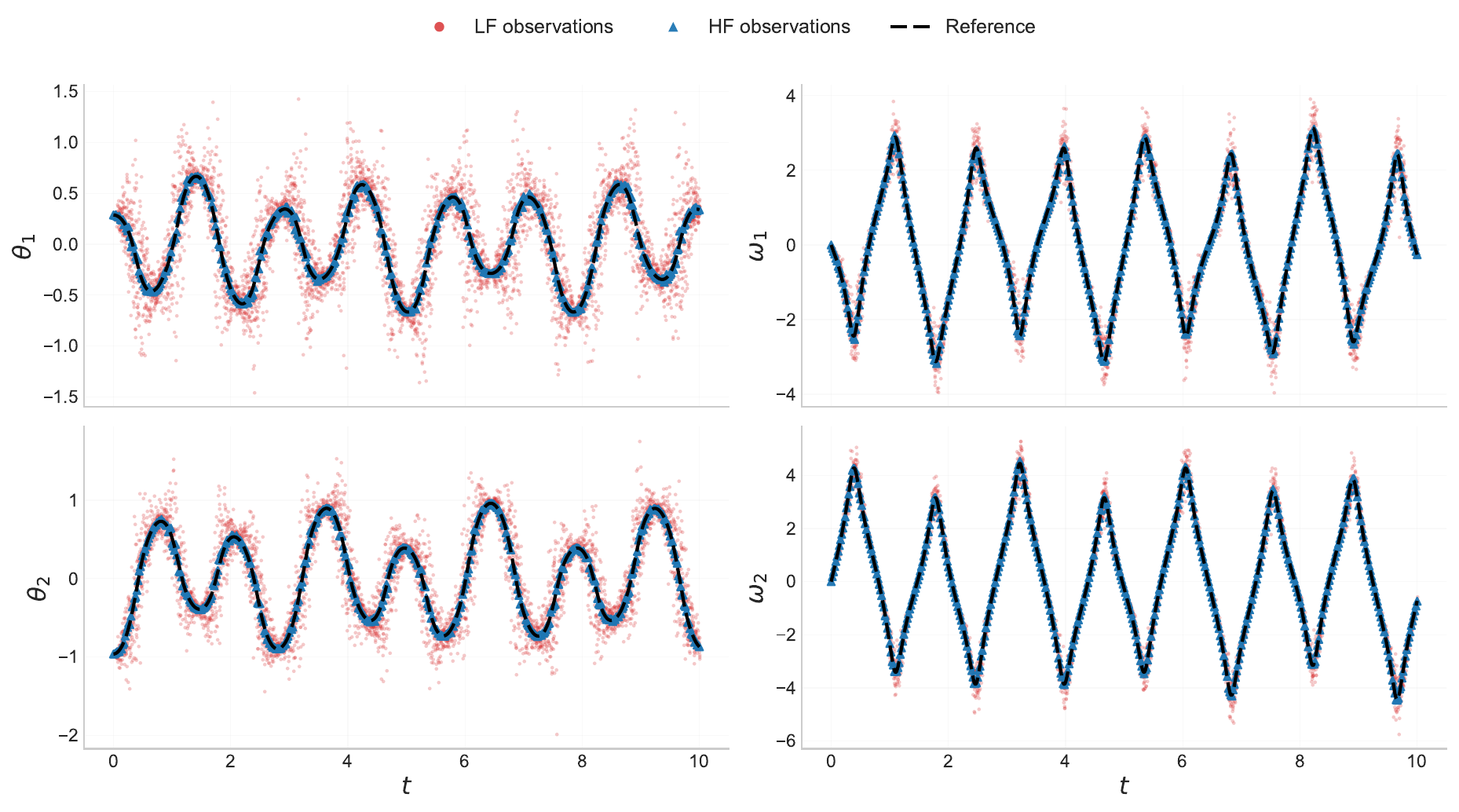}
    \vspace{-0.45cm}
    \caption{Observed double-pendulum states over time. Dashed black lines denote the reference trajectory; blue and red markers denote HF and LF observations, respectively.}
    \label{fig:double-pendulum-observations}
    \vspace{-0.4cm}
\end{figure}
\begin{figure}[t]
    \centering
    \includegraphics[width=\textwidth]{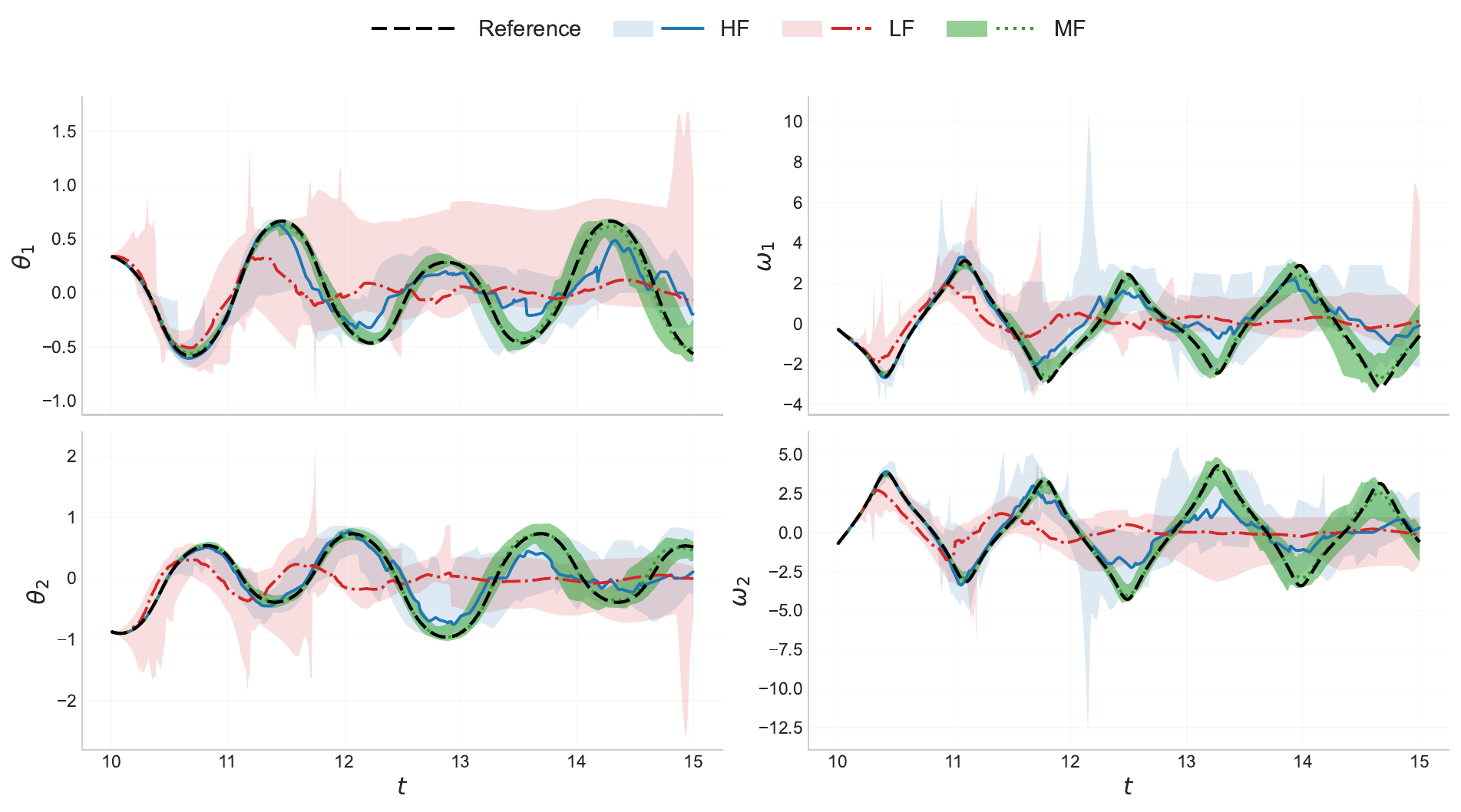}
    \vspace{-0.45cm}
    \caption{Double-pendulum forecasting with a degree-3 polynomial library. Dashed black lines denote the reference future trajectory. Colored lines and bands denote ensemble median forecasts and the associated \(90\%\) confidence intervals for the three forecasting settings: high-fidelity data only, low-fidelity data only, and multi-fidelity fusion.}
    \label{fig:double-pendulum-forecast}
    \vspace{-0.4cm}
\end{figure}
\subsection{Part III: Forecasting Benchmark on the Double Pendulum}\label{subsec:res3}
Part~III investigates the forecasting capabilities of MF-SINDy in terms of accuracy and uncertainty. In this setting, the objective is not exact recovery of the governing equations, but rather building a cheap and interpretable surrogate of a dynamical system. The benchmark problem we consider is the planar double pendulum with state variables \(\theta_1,\theta_2\) and corresponding angular velocities \(\omega_1=\dot{\theta}_1\), \(\omega_2=\dot{\theta}_2\), see Figure \ref{fig:double-pendulum-setup}. Assuming unit masses and unit rod lengths and gravity acceleration $g$, the dynamics can be written in second-order form as a coupled system for \(\theta_1\) and \(\theta_2\), as follows:
\begin{equation}
\frac{1}{3}\ddot{\theta}_2
+\frac{1}{2}\ddot{\theta}_1\cos(\theta_1-\theta_2)
-\frac{1}{2}\dot{\theta}_1^{\,2}\sin(\theta_1-\theta_2)
+\frac{1}{2}g\sin\theta_2 = 0,
\end{equation}
with an analogous companion equation for \(\theta_1\). Rewriting this model as a first-order dynamical system in the state $\mathbf{u}(t)=\bigl(\theta_1(t),\theta_2(t),\omega_1(t),\omega_2(t)\bigr) $
leads to a considerably more cumbersome set of coupled trigonometric and rational expressions. For this reason, we adopt a simpler surrogate representation based on a polynomial candidate library. The polynomial degree is selected by grid search over the range \(1\) to \(5\), and degree \(3\) is retained for all reported experiments.

The reference trajectories are generated from random initial conditions satisfying
$
\theta_1(0),\theta_2(0)\in[-\pi/3,\pi/3],$ and $
\omega_1(0),\omega_2(0)\in[-1,1].
$
This choice avoids excessively energetic configurations while still yielding a dynamically rich and challenging forecasting regime.

Unlike Parts I and II, the HF and LF data in this case are not obtained from trajectories with different initial conditions, but from repeated noisy observations of a single underlying trajectory. In the benchmark implementation, one HF realization and five LF realizations are sampled from a common reference path, with a time-dependent observation variance proportional to $\|(\omega_1(t),\omega_2(t))\|$ and different proportionality coefficient:  0.01 for HF observations and 0.1 for LF ones. As shown in Figure \ref{fig:double-pendulum-setup}, this leads to a clear separation in dispersion between fidelity levels.

Figure~\ref{fig:double-pendulum-observations} further shows that this dispersion is state dependent. The LF observations spread most during phases of larger angular-speed magnitude, consistent with the velocity-dependent noise model, whereas the HF observations remain comparatively concentrated throughout the observation window.

We compare three covariance-weighted weak-SINDy forecasting settings: one trained using only high-fidelity observations, one trained using only low-fidelity observations, and one trained by combining both fidelity groups in a single weighted weak-regression problem. In all three settings, covariance-aware weighting is applied under the assumption that the observation variances are known a priori, together with weak SINDy and ensemble SINDy.

Predictive performance is quantified using three complementary metrics. Accuracy is measured by the root mean squared error (RMSE) between the reference trajectory and the ensemble median forecast. Uncertainty sharpness is summarized by the mean width of the \(90\%\) confidence intervals. The meaningfulness of the uncertainty intervals is assessed through their empirical coverage, namely the proportion of reference states contained within the corresponding intervals over the forecasting horizon.

Figure~\ref{fig:double-pendulum-forecast} indicates that combining the two fidelity levels provides the best trade-off between bias and dispersion. The model identified from low-fidelity observations alone is the least accurate and produces the widest forecast bands. The model identified from high-fidelity observations alone often follows the reference trajectory more closely, but its predictions remain more variable because it is fitted from much less data. By combining the cleaner high-fidelity observation with repeated low-fidelity measurements, the multi-fidelity model yields forecast medians that remain closer to the reference trajectory over most of the prediction horizon while maintaining comparatively tight uncertainty bands. 
\begin{table}[t]
\centering
\caption{Forecasting metrics for the double-pendulum benchmark, averaged over 25 random seeds. Width denotes the mean of the width of the \(90\%\) confidence interval. Values are reported as mean \(\pm\) standard deviation over seeds.}
\label{tab:double-pendulum-forecast}
\begin{tabular}{lccc}
\hline
Model  & RMSE & Width & Coverage \\
\hline
High-fidelity only & $0.669 \pm 0.419$ & $1.117 \pm 0.6460 $ & $0.703 \pm 0.2460$ \\
Low-fidelity only & $0.841 \pm 0.290$ & $1.718 \pm 1.0382$ & $0.660 \pm 0.1629$ \\
Multi-fidelity fusion & $\mathbf{0.279 \pm 0.231}$ & $\mathbf{0.801 \pm 0.3552}$ & $\mathbf{0.877 \pm 0.0994}$ \\
\hline
\end{tabular}
\vspace{-0.1cm}
\end{table}
The quantitative results in Table~\ref{tab:double-pendulum-forecast} confirm the visual trends in Figure~\ref{fig:double-pendulum-forecast}. Across 25 random seeds, the multi-fidelity model reduces RMSE by about 58\% relative to the high-fidelity-only model and 67\% relative to the low-fidelity-only model, while producing predictive bands that are narrower. It also achieves the strongest empirical coverage, indicating improved uncertainty quantification despite library misspecification.

\section{Discussion}\label{sec:discussion}

Our results consistently show that weighting allows for an effective integration of heterogeneous data. In all the considered settings, covariance-aware weighting improves sparse recovery relative to unweighted alternatives. In the forecasting benchmark, the same idea leads to more accurate predictions and more informative uncertainty bands, even under deliberate library misspecification. Hence, the benefit of multi-fidelity data does not come simply from having more observations, but from combining them in a way that reflects their  reliability. 

Assembling and rescaling the covariance matrices provide an almost negligible additional cost to the overall procedure, so that when the variance profile is known, their usage is strongly recommended. Moreover, a practical takeaway is that best performances are obtained when the support of the weak test functions remains relatively small: overly large supports reduce local resolution and may blur distinctions between cleaner and noisier regions of the data.   
Compared with approaches such as \cite{conti_mf}, the proposed procedure identifies an interpretable surrogate model and uses multi-fidelity data only during the offline training stage. As a result, the learned model can be deployed for forecasting even when low-fidelity data are not available at inference time. In contrast to \cite{MENG2025113651}, multi-fidelity information is incorporated directly into the sparse identification step rather than through an intermediate surrogate used to reconstruct trajectories. This avoids the additional modeling layer and the hyperparameter-tuning burden associated with multi-fidelity Gaussian processes. Noise mitigation is instead handled naturally by the weak \gls{sindy} formulation itself, which is generally characterized by a simpler tuning procedure. At the same time, the present framework relies on the assumption that the residuals are unbiased, which is not the case when relying on the multi fidelity GP.

At the same time, the study also contains several limitations. The covariance model is derived at leading order and neglects the perturbation induced in the weak feature matrix, which may become important in more challenging PDE settings or at higher noise levels. In addition, the present benchmarks assume that all fidelities share the same underlying dynamics and differ mainly through their noise level. 
These limitations point to several directions for further extensions, for instance by considering {\em (i)} more refined covariance models that include library matrix perturbation, {\em (ii)} adaptive estimation of variance profiles from data, and {\em (iii)} situations in which fidelity differences involve not only noise magnitude, but also systematic model discrepancy. It would also be valuable to study adaptive choices of weak test-function support and overlap in order to better balance smoothing and local resolution.

\section{Conclusion}\label{sec:conclusion}

This work introduced MF-SINDy, a covariance-aware multi-fidelity extension of weak and ensemble \gls{sindy} for sparse dynamical-system identification from heterogeneous data. The proposed framework combines weak regression, ensemble-based model discovery, and generalized least-squares weighting in order to account for differences in measurement reliability both across trajectories and within individual trajectories. The numerical experiments consistently show that covariance-aware weighting improves sparse model identification in both the multi-trajectory and heteroscedastic single-trajectory settings, investigating the performance on several dynamical systems including ODEs and PDEs. In the forecasting benchmark, the same strategy also improves predictive accuracy and uncertainty quantification. Indeed, heterogeneous data can be effectively exploited when their reliability is incorporated directly into the sparse regression problem, improving accuracy and reducing uncertainty in both recovered coefficients and forecasts.

Overall, MF-SINDy provides an interpretable and statistically principled extension of sparse equation discovery to multi-fidelity and nonuniform-noise settings. The results suggest that combining weak formulations with covariance-aware weighting is a promising direction for robust system identification from heterogeneous observations. Future work will focus on richer covariance models, adaptive estimation of variance profiles, and extensions to settings in which fidelity differences also involve systematic bias or partial model discrepancy.

\paragraph{Code Availability}
The code for implementing MF-SINDy and reproduce the results of the paper is available at \url{https://github.com/filippozacchei/MFSindy} alongside guiding tutorials.

\paragraph{Acknowledgements}

FZ acknowledges the support of the JRC STEAM STM-Politecnico di Milano agreement. AM acknowledges the project “Dipartimento di Eccellenza” 2023–2027 funded by MUR, and the Project “Reduced Order Modeling and Deep Learning for the real-time approximation of PDEs (DREAM)” (Starting Grant No. FIS00003154), funded by the Italian Science Fund (FIS) - Ministero dell’Università e della Ricerca. ALJ and SLB acknowledge support from the National Science Foundation AI Institute in Dynamic Systems (grant number 2112085) and the Boeing Company.

\enlargethispage{20pt}
\appendix

\section{Justification of the weak-space covariance approximation}
\label{app:covariance_approx}

This appendix justifies the covariance model used in the main text. The goal is to explain why, in the weak formulation, the dominant stochastic contribution to the residual can be approximated by the perturbation entering through the weak left-hand side, while the perturbation induced in the weak feature matrix is neglected at leading order. The discussion is first carried out in the ODE setting, where the argument is most transparent, and then briefly revisited for multiple trajectories and for PDEs.

The key idea is the same as in the residual analysis underlying weak SINDy: once the governing equation is written in weak form and evaluated on noisy data, the residual at the target coefficients contains a contribution coming from the weak left-hand side, a contribution coming from perturbing the library evaluation, and a deterministic quadrature error. In the ODE setting, the first of these terms provides the leading covariance model used in the main text. In the PDE setting, the same construction remains valid formally, but the dominance of that term is more delicate and requires an additional support-scaling assumption.

\subsection{Residual decomposition in the ODE setting}

Consider a single trajectory and write $\widehat{\mathbf{U}}=\mathbf{U}+\mathbf{E}$, where \(\widehat{\mathbf{U}}\) is the noisy trajectory, \(\mathbf{U}\) is the corresponding noiseless trajectory, and \(\mathbf{E}\) collects the measurement perturbations. The weakly assembled quantities are
\begin{equation}
    \mathbf{b}=-\dot{\textbf{V}}\widehat{\mathbf{U}},
    \qquad
    \mathbf{G}=\mathbf{V}\,\boldsymbol{\Theta}(\widehat{\mathbf{U}}),
    \qquad
    \mathbf{R}(\boldsymbol{\Xi})=\mathbf{b}-\mathbf{G}\boldsymbol{\Xi}.
\end{equation}
To characterize the covariance used in the main text, we evaluate the residual at the target coefficient matrix \(\boldsymbol{\Xi}^\star\):
\begin{equation}
    \mathbf{R}(\boldsymbol{\Xi}^\star)
    =
    -\dot{\textbf{V}}\widehat{\mathbf{U}}
    -
    \mathbf{V}\boldsymbol{\Theta}(\widehat{\mathbf{U}})\boldsymbol{\Xi}^\star.
    \label{eq:app_resid_start}
\end{equation}
Now let \(\mathbf{I}_q\) denote the deterministic quadrature remainder obtained when the weak identity is evaluated along the noiseless trajectory. Assume than that $\boldsymbol{\Theta}(\mathbf{U})$ is expressive enough to capture the true model underlying the dynamical system. Then:
\begin{equation}
    -\dot{\textbf{V}}\mathbf{U}
    =
    \mathbf{V}\boldsymbol{\Theta}(\mathbf{U})\boldsymbol{\Xi}^\star
    + \mathbf{I}_q.
    \label{eq:app_true_weak_relation}
\end{equation}
Substituting \(\widehat{\mathbf{U}}=\mathbf{U}+\mathbf{E}\) into \eqref{eq:app_resid_start} and using \eqref{eq:app_true_weak_relation}, we obtain
\begin{equation}
    \mathbf{R}(\boldsymbol{\Xi}^\star)
    =
    -\dot{\textbf{V}}\mathbf{E}
    +
    \mathbf{V}\Bigl(\boldsymbol{\Theta}(\mathbf{U})-\boldsymbol{\Theta}(\widehat{\mathbf{U}})\Bigr)\boldsymbol{\Xi}^\star
    +
    \mathbf{I}_q.
    \label{eq:app_resid_exact}
\end{equation}
Thus, the residual at \(\boldsymbol{\Xi}^\star\) contains three contributions: a perturbation entering through the weak left-hand side, a perturbation induced by evaluating the library on noisy data, and a deterministic quadrature remainder.

To make the second term more explicit, assume that each candidate function defining the library map \(\boldsymbol{\Theta}\) is \(C^1\) on a neighborhood of the trajectory range. For each sampling time \(t_i\), let
\[
\boldsymbol{\Theta}(\mathbf{u}_i)
=
\bigl[\theta_1(\mathbf{u}_i),\ldots,\theta_D(\mathbf{u}_i)\bigr]
\in\mathbb{R}^{1\times D},
\]
and denote by
\[
J_{\Theta}(\mathbf{u}_i)
=
\begin{bmatrix}
\nabla \theta_1(\mathbf{u}_i)^\top\\
\vdots\\
\nabla \theta_D(\mathbf{u}_i)^\top
\end{bmatrix}
\in\mathbb{R}^{D\times d}
\]
the Jacobian of the library map at \(\mathbf{u}_i\), whose \(j\)th row is the gradient of the \(j\)th candidate function evaluated at \(\mathbf{u}_i\). A first-order Taylor expansion then gives
\begin{equation}
    \boldsymbol{\Theta}(\widehat{\mathbf{u}}_i)
    =
    \boldsymbol{\Theta}(\mathbf{u}_i)
    +
    \bigl(J_{\Theta}(\mathbf{u}_i)\,\boldsymbol{\varepsilon}_i\bigr)^\top
    +
    \boldsymbol{\rho}_i,
\end{equation}
where \(\boldsymbol{\rho}_i\in\mathbb{R}^{1\times D}\) satisfies
\[
\|\boldsymbol{\rho}_i\|_2=\mathcal{O}(\|\boldsymbol{\varepsilon}_i\|_2^2).
\]
Stacking these expansions row-wise yields
\begin{equation}
    \boldsymbol{\Theta}(\widehat{\mathbf{U}})
    =
    \boldsymbol{\Theta}(\mathbf{U})
    +
    \mathbf{J}_{\Theta}(\mathbf{U};\mathbf{E})
    +
    \mathbf{H}_{\Theta}(\mathbf{U};\mathbf{E}),
    \label{eq:app_theta_expand}
\end{equation}
where the \(i\)th row of \(\mathbf{J}_{\Theta}(\mathbf{U};\mathbf{E})\in\mathbb{R}^{N\times D}\) is
$\bigl(J_{\Theta}(\mathbf{u}_i)\,\boldsymbol{\varepsilon}_i\bigr)^\top$, and \(\mathbf{H}_{\Theta}(\mathbf{U};\mathbf{E})\) collects the higher-order remainders, with
\[
\|\mathbf{H}_{\Theta}(\mathbf{U};\mathbf{E})\|_F
=
\mathcal{O}(\|\mathbf{E}\|_F^2).
\]
Substituting \eqref{eq:app_theta_expand} into \eqref{eq:app_resid_exact} yields
\begin{equation}
    \mathbf{R}(\boldsymbol{\Xi}^\star)
    =
    -\dot{\textbf{V}}\mathbf{E}
    -
    \mathbf{V}\mathbf{J}_{\Theta}(\mathbf{U};\mathbf{E})\boldsymbol{\Xi}^\star
    +
    \mathbf{I}_q
    +
    \mathcal{O}(\|\mathbf{E}\|_F^2).
    \label{eq:app_resid_firstorder}
\end{equation}
Thus, up to the deterministic quadrature remainder, the weak residual contains two first-order stochastic contributions: a derivative-driven term and a feature-driven term.

\subsection{Leading-order covariance approximation}

Neglecting the deterministic quadrature remainder \(\mathbf{I}_q\), assumed small under sufficiently accurate quadrature, together with the higher-order term \(\mathcal{O}(\|\mathbf{E}\|_F^2)\), the residual evaluated at \(\boldsymbol{\Xi}^\star\) is approximated at first order by
\begin{equation}
    \mathbf{R}(\boldsymbol{\Xi}^\star)
    \approx
    -\dot{\textbf{V}}\mathbf{E}
    -
    \mathbf{V}\mathbf{J}_{\Theta}(\mathbf{U};\mathbf{E})\boldsymbol{\Xi}^\star.
    \label{eq:app_resid_firstorder_reduced}
\end{equation}
Thus, the weak residual contains two first-order stochastic contributions: a derivative-driven term entering through the weak left-hand side, and a feature-driven term induced by evaluating the library on noisy data.
The covariance model used in the main text is obtained by retaining only the derivative-driven contribution in \eqref{eq:app_resid_firstorder_reduced}. This amounts to treating the perturbation of the weak feature matrix as negligible at leading order, so that the generalized least-squares weighting is determined by the covariance induced by the left-hand side alone. Accordingly, for each component \(r\),
\begin{equation}
    \mathrm{Cov}\!\bigl(\mathbf{R}_{\cdot r}(\boldsymbol{\Xi}^\star)\bigr)
    \approx
    \mathrm{Cov}\!\bigl((-\dot{\textbf{V}}\mathbf{E})_{\cdot r}\bigr).
    \label{eq:app_cov_leading}
\end{equation}
In the heteroscedastic setting, where
\begin{equation}
    \mathrm{Cov}(\boldsymbol{\varepsilon}_i)=\sigma_i^2\mathbf{I}_d,
    \qquad
    \mathbf{D}_\sigma=\mathrm{diag}(\sigma_1^2,\ldots,\sigma_N^2),
\end{equation}
this yields
\begin{equation}
    \mathrm{Cov}\!\bigl(\mathbf{R}_{\cdot r}(\boldsymbol{\Xi}^\star)\bigr)
    \approx
    \dot{\textbf{V}}\mathbf{D}_\sigma(\dot{\textbf{V}})^\top.
    \label{eq:app_hetero_cov}
\end{equation}
To justify \eqref{eq:app_hetero_cov}, it is enough to show that the feature-driven term is asymptotically smaller than the derivative-driven one. This follows from the different scaling of \(\mathbf{V}\) and \(\dot{\textbf{V}}\) with respect to the support width \(h\) of the test functions. In the weak formulation used here, the test functions are compactly supported polynomial weights obtained by rescaling a fixed profile, for example
\[
\varphi_{k,h}(t)=\left(1-\left(\frac{t-t_k}{h}\right)^2\right)^p
\]
on its support. For such localized test functions, one has the standard scalings
\begin{equation}
    \|\mathbf{V}\|_2^2=\mathcal{O}(h),
    \qquad
    \|\dot{\textbf{V}}\|_2^2=\mathcal{O}(h^{-1}).
\end{equation}

To see that the Jacobian term does not alter this scaling, assume that the noiseless trajectory remains in a compact set \(K\subset\mathbb{R}^d\), that each candidate function in the library is \(C^1\) on a neighborhood of \(K\). Then the Jacobian of the library map is uniformly bounded along the trajectory, i.e.
\[
\sup_{i=1,\dots,N}\|J_\Theta(\mathbf{u}_i)\|_2 \le C_J.
\]
Consequently,
\[
\|J_\Theta(\mathbf{u}_i)\boldsymbol{\varepsilon}_i\|_2
\le
C_J \|\boldsymbol{\varepsilon}_i\|_2,
\]
and therefore
\[
\|\mathbf{V}\mathbf{J}_\Theta(\mathbf{U};\mathbf{E})\boldsymbol{\Xi}^\star\|_F
\lesssim
\|\mathbf{V}\|_2\,\|\mathbf{E}\|_F.
\]
Thus, the feature-driven contribution therefore inherits the \(\mathcal{O}(h)\) scaling of \(\mathbf{V}\), while the derivative-driven contribution inherits the \(\mathcal{O}(h^{-1})\) scaling of \(\dot{\textbf{V}}\). Hence their ratio scales as
\begin{equation}
    \frac{\mathcal{O}(h)}{\mathcal{O}(h^{-1})}
    =
    \mathcal{O}(h^2)\to 0
    \qquad \text{as } h\to 0.
\end{equation}
Therefore, for sufficiently localized temporal test functions, the perturbation entering through the weak left-hand side dominates the perturbation induced by the library evaluation, and the leading-order covariance of the weak residual is well approximated by \eqref{eq:app_hetero_cov}.

This argument also highlights a practical tradeoff in the choice of the test-function support width. On the one hand, smaller supports favor the asymptotic separation above, since they increase the relative dominance of the derivative-driven term over the feature-driven one. On the other hand, larger supports provide stronger averaging and therefore more effective noise smoothing in the weak formulation. In practice, the support width must balance these two effects: it should be small enough for the covariance approximation to remain accurate, but not so small that the weak formulation loses its denoising advantage.

\subsection{Remark on the PDE setting}

The same covariance argument extends formally to the PDE setting. 
As in the ODE case, the covariance model used in the main text is obtained by retaining only the leading stochastic contribution entering through the weak temporal left-hand side. 
Unlike in the ODE setting, however, the perturbation induced in the weak feature matrix is not automatically negligible. After integration by parts, spatial derivatives in the library are transferred onto the test functions, so the feature-driven term may involve factors such as \(\partial_{x_j}\phi\) or \(\partial_{x_jx_j}\phi\), whose magnitude increases as the spatial support shrinks. If the temporal support has width \(h_t\) and the spatial supports have widths \(h_{x_1},\dots,h_{x_s}\), then the temporal derivative-driven contribution scales like
\[
\mathcal{O}\!\left(\frac{\prod_{j=1}^s h_{x_j}}{h_t}\right),
\]
whereas a feature-driven contribution involving an \(m\)-th spatial derivative in direction \(x_\ell\) scales like
\[
\mathcal{O}\!\left(
h_t\,h_{x_\ell}^{\,1-2m}\prod_{j\neq \ell} h_{x_j}
\right).
\]
Hence their ratio behaves as
\[
\mathcal{O}\!\left(h_t^2\,h_{x_\ell}^{-2m}\right).
\]
Therefore, the approximation \eqref{eq:app_hetero_cov} is justified only under an additional scale-separation assumption, namely that the temporal support shrinks sufficiently faster than the relevant spatial supports. For example, for second-order spatial derivatives it is sufficient to require
\[
h_t^2 h_{x_\ell}^{-4}\to 0.
\]

Under such a regime, the perturbation entering through the weak temporal left-hand side remains dominant and \eqref{eq:app_hetero_cov} provides the appropriate leading-order covariance model. If this condition is not satisfied, then the perturbation of \(\mathbf{G}\) is no longer negligible, and the generalized least-squares approximation based only on \eqref{eq:app_hetero_cov} becomes incomplete. In that case, one must either choose test functions with stronger spatial averaging or include the feature-driven contribution explicitly in the covariance model.

\bibliographystyle{unsrt}
\bibliography{references}

@book{brunton2019, 
title={Data-Driven Science and Engineering: Machine Learning, Dynamical Systems, and Control}, 
publisher={Cambridge University Press}, 
author={Brunton, Steven L. and Kutz, J. Nathan}, 
year={2019}}

@book{chandrupatla2021, place={Cambridge}, edition={5}, title={Introduction to Finite Elements in Engineering}, publisher={Cambridge University Press}, author={Chandrupatla, Tirupathi and Belegundu, Ashok}, year={2021}}

@book{quarteroni2016,
	title = {Reduced {Basis} {Methods} for {Partial} {Differential} {Equations}},
	volume = {92},
	publisher = {Springer International Publishing},
	author = {Quarteroni, Alfio and Manzoni, Andrea and Negri, Federico},
	year = {2016},
	doi = {10.1007/978-3-319-15431-2},
}

@book{forrester2008,
	edition = {1},
	title = {Engineering {Design} via {Surrogate} {Modelling}: {A} {Practical} {Guide}},
	publisher = {Wiley},
	author = {Forrester, Alexander I. J. and Sóbester, András and Keane, Andy J.},
	month = jul,
	year = {2008},
	doi = {10.1002/9780470770801},
}

@book{greene2002econometricanalysis,
  author    = {Greene, William H.},
  title     = {Econometric Analysis},
  edition   = {5},
  year      = {2003},
  publisher = {Prentice Hall},
  isbn      = {9780130661890}
}

@article{tao2024,
  title={Advancements and challenges of digital twins in industry},
  author={Tao, Fei and Zhang, He and Zhang, Chenyuan},
  journal={Nature Computational Science},
  volume={4},
  number={3},
  pages={169--177},
  year={2024},
  doi={10.1038/s43588-024-00603-w}
}

@article{kapteyn2021,
	title = {A probabilistic graphical model foundation for enabling predictive digital twins at scale},
	volume = {1},
	doi = {10.1038/s43588-021-00069-0},
	number = {5},
	journal = {Nature Computational Science},
	author = {Kapteyn, Michael G. and Pretorius, Jacob V. R. and Willcox, Karen E.},
	month = may,
	year = {2021},
	pages = {337--347},
}

@article{torzoni2024,
	title = {A digital twin framework for civil engineering structures},
	volume = {418},
	doi = {10.1016/j.cma.2023.116584},
	journal = {Computer Methods in Applied Mechanics and Engineering},
	author = {Torzoni, Matteo and Tezzele, Marco and Mariani, Stefano and Manzoni, Andrea and Willcox, Karen E.},
	month = jan,
	year = {2024},
	pages = {116584},
}

@article{mcclellan2022,
	title = {A physics-based digital twin for model predictive control of autonomous unmanned aerial vehicle landing},
	volume = {380},
	doi = {10.1098/rsta.2021.0204},
	number = {2229},
	journal = {Philosophical Transactions of the Royal Society A: Mathematical, Physical and Engineering Sciences},
	author = {McClellan, Andrew and Lorenzetti, Joseph and Pavone, Marco and Farhat, Charbel},
	month = jun,
	year = {2022},
}

@article{benner2015,
	title = {A {Survey} of {Projection}-{Based} {Model} {Reduction} {Methods} for {Parametric} {Dynamical} {Systems}},
	volume = {57},
	doi = {10.1137/130932715},
	number = {4},
	journal = {SIAM Review},
	author = {Benner, Peter and Gugercin, Serkan and Willcox, Karen},
	month = jan,
	year = {2015},
	pages = {483--531},
}

@article{samadian2025,
	title = {Application of {Data}-{Driven} {Surrogate} {Models} in {Structural} {Engineering}: {A} {Literature} {Review}},
	volume = {32},
	doi = {10.1007/s11831-024-10152-0},
	number = {2},
	journal = {Archives of Computational Methods in Engineering},
	author = {Samadian, Delbaz and Muhit, Imrose B. and Dawood, Nashwan},
	month = mar,
	year = {2025},
	pages = {735--784},
}

@article{brunton2016,
    doi = {10.1073/pnas.1517384113},
    title={Discovering governing equations from data by sparse identification of nonlinear dynamical systems},
    author={Brunton, Steven L and Proctor, Joshua L and Kutz, J Nathan},
    journal={Proceedings of the National Academy of Sciences},
    volume={113},
    number={15},
    pages={3932--3937},
    year={2016},
}

@article{brunton2020,
	title = {Machine {Learning} for {Fluid} {Mechanics}},
	volume = {52},
	doi = {10.1146/annurev-fluid-010719-060214},
	language = {en},
	number = {Volume 52, 2020},
	journal = {Annual Review of Fluid Mechanics},
	author = {Brunton, Steven L. and Noack, Bernd R. and Koumoutsakos, Petros},
	month = jan,
	year = {2020},
	pages = {477--508},
}

@article{peherstorfer2018,
	title = {Survey of {Multifidelity} {Methods} in {Uncertainty} {Propagation}, {Inference}, and {Optimization}},
	volume = {60},
	doi = {10.1137/16M1082469},
	number = {3},
	journal = {SIAM Review},
	author = {Peherstorfer, Benjamin and Willcox, Karen and Gunzburger, Max},
	month = jan,
	year = {2018},
	pages = {550--591},
}

@article{kaiser2018,
	title = {Sparse identification of nonlinear dynamics for model predictive control in the low-data limit},
	volume = {474},
	doi = {10.1098/rspa.2018.0335},
	number = {2219},
	journal = {Proceedings of the Royal Society A: Mathematical, Physical and Engineering Sciences},
	author = {Kaiser, E. and Kutz, J. N. and Brunton, S. L.},
	month = nov,
	year = {2018},
	pages = {20180335},
}

@article{fasel2022,
	title = {Ensemble-{SINDy}: {Robust} sparse model discovery in the low-data, high-noise limit, with active learning and control},
	volume = {478},
	doi = {10.1098/rspa.2021.0904},
	journal = {Proceedings of the Royal Society A: Mathematical, Physical and Engineering Sciences},
	author = {Fasel, U. and Kutz, J. N. and Brunton, B. W. and Brunton, S. L.},
	month = apr,
	year = {2022},
	pages = {20210904},
}

@article{messenger2021_1,
	title = {Weak {SINDy}: {Galerkin}-{Based} {Data}-{Driven} {Model} {Selection}},
	volume = {19},
	doi = {10.1137/20M1343166},
	number = {3},
	journal = {Multiscale Modeling \& Simulation},
	author = {Messenger, Daniel A. and Bortz, David M.},
	month = jan,
	year = {2021},
	pages = {1474--1497},
}

@article{messenger2021_2,
	title = {Weak {SINDy} for partial differential equations},
	volume = {443},
	doi = {10.1016/j.jcp.2021.110525},
	journal = {Journal of Computational Physics},
	author = {Messenger, Daniel A. and Bortz, David M.},
	month = oct,
	year = {2021},
	pages = {110525},
}

@article{rudy2017,
    author = {Samuel H. Rudy  and Steven L. Brunton  and Joshua L. Proctor  and J. Nathan Kutz },
    title = {Data-driven discovery of partial differential equations},
    journal = {Science Advances},
    volume = {3},
    number = {4},
    pages = {e1602614},
    year = {2017},
    doi = {10.1126/sciadv.1602614},
}

@article{kaptanoglu2021,
  title = {Promoting global stability in data-driven models of quadratic nonlinear dynamics},
  author = {Kaptanoglu, Alan A. and Callaham, Jared L. and Aravkin, Aleksandr and Hansen, Christopher J. and Brunton, Steven L.},
  journal = {Physical Review Fluids},
  volume = {6},
  issue = {9},
  pages = {094401},
  numpages = {30},
  year = {2021},
  month = {Sep},
  doi = {10.1103/PhysRevFluids.6.094401},
}

@article{loiseau2018, 
    title={Sparse reduced-order modelling: sensor-based dynamics to full-state estimation}, 
    volume={844}, 
    doi={10.1017/jfm.2018.147}, 
    journal={Journal of Fluid Mechanics}, 
    author={Loiseau, Jean-Christophe and Noack, Bernd R. and Brunton, Steven L.}, 
    year={2018}, 
    pages={459–490}
}

@article{schaeffer2017,
  title = {Sparse model selection via integral terms},
  author = {Schaeffer, Hayden and McCalla, Scott G.},
  journal = {Phys. Rev. E},
  volume = {96},
  issue = {2},
  pages = {023302},
  numpages = {7},
  year = {2017},
  month = {Aug},
  doi = {10.1103/PhysRevE.96.023302},
}

@article{charnes_equivalence_1976,
	title = {The {Equivalence} of {Generalized} {Least} {Squares} and {Maximum} {Likelihood} {Estimates} in the {Exponential} {Family}},
	volume = {71},
	issn = {0162-1459},
	doi = {10.1080/01621459.1976.10481508},
	number = {353},
	journal = {Journal of the American Statistical Association},
	author = {Charnes, A. and Frome, E. L. and Yu, P. L.},
	month = mar,
	year = {1976},
	pages = {169--171},
}

@article{shaffer_gauss_markov_1991,
    author = {Juliet Popper Shaffer},
    title = {The Gauss—Markov Theorem and Random Regressors},
    journal = {The American Statistician},
    volume = {45},
    number = {4},
    pages = {269--273},
    year = {1991},
    publisher = {Taylor \& Francis},
    doi = {10.1080/00031305.1991.10475819},
}

@article{aitken1936,
  author  = {Aitken, A. C.},
  title   = {IV.—On Least Squares and Linear Combination of Observations},
  journal = {Proceedings of the Royal Society of Edinburgh},
  volume  = {55},
  pages   = {42--48},
  year    = {1936},
  doi     = {10.1017/S0370164600014346}
}

@article{forrester2007multi,
    author = {Forrester, Alexander I.J and Sóbester, András and Keane, Andy J},
    title = {Multi-fidelity optimization via surrogate modelling},
    journal = {Proceedings of the Royal Society A: Mathematical, Physical and Engineering Sciences},
    volume = {463},
    number = {2088},
    pages = {3251-3269},
    year = {2007},
    month = {10},
    doi = {10.1098/rspa.2007.1900},
}

@article{perdikaris1,
    author = {Perdikaris, P. and Venturi, D. and Royset, J. O. and Karniadakis, G. E.},
    title = {Multi-fidelity modelling via recursive co-kriging and Gaussian–Markov random fields},
    journal = {Proceedings of the Royal Society A: Mathematical, Physical and Engineering Sciences},
    volume = {471},
    number = {2179},
    pages = {20150018},
    year = {2015},
    month = {07},
    doi = {10.1098/rspa.2015.0018},  
}

@article{BRYSON2017121,
    title = {All-at-once approach to multifidelity polynomial chaos expansion surrogate modeling},
    journal = {Aerospace Science and Technology},
    volume = {70},
    pages = {121-136},
    year = {2017},
    issn = {1270-9638},
    doi = {10.1016/j.ast.2017.07.043},
    author = {Dean E. Bryson and Markus P. Rumpfkeil},

}

@article{CHENG2019360,
    title = {Multi-level multi-fidelity sparse polynomial chaos expansion based on Gaussian process regression},
    journal = {Computer Methods in Applied Mechanics and Engineering},
    volume = {349},
    pages = {360-377},
    year = {2019},
    doi = {10.1016/j.cma.2019.02.021},
    author = {Kai Cheng and Zhenzhou Lu and Ying Zhen},
}

@article{Du,
    author = {Du, Xiaosong and Leifsson, Leifur},
    year = {2020},
    month = {01},
    pages = {},
    title = {Multifidelity Modeling by Polynomial Chaos-Based Cokriging to Enable Efficient Model-Based Reliability Analysis of NDT Systems},
    volume = {39},
    journal = {Journal of Nondestructive Evaluation},
    doi = {10.1007/s10921-020-0656-8}
}

@article{le2014recursive,
    title={Recursive co-kriging model for design of computer experiments with multiple levels of fidelity},
    author={Le Gratiet, Loic and Garnier, Josselin},
    journal={International Journal for Uncertainty Quantification},
    volume={4},
    number={5},
    year={2014},
    doi={10.1615/Int.J.UncertaintyQuantification.2014006914},
    pages={365-386}
}

@article{MENG2020109020,
    title = {A composite neural network that learns from multi-fidelity data: Application to function approximation and inverse PDE problems},
    journal = {Journal of Computational Physics},
    volume = {401},
    pages = {109020},
    year = {2020},
    doi = {10.1016/j.jcp.2019.109020},
    author = {Xuhui Meng and George Em Karniadakis},
}

@article{PENWARDEN2022110844,
    title = {Multifidelity modeling for Physics-Informed Neural Networks (PINNs)},
    journal = {Journal of Computational Physics},
    volume = {451},
    pages = {110844},
    year = {2022},
    doi = {10.1016/j.jcp.2021.110844},
    author = {Michael Penwarden and Shandian Zhe and Akil Narayan and Robert M. Kirby},
}

@misc{niu24d,
    title={Multi-Fidelity Residual Neural Processes for Scalable Surrogate Modeling}, 
    author={Ruijia Niu and Dongxia Wu and Kai Kim and Yi-An Ma and Duncan Watson-Parris and Rose Yu},
    year={2024},
    eprint={2402.18846},
    archivePrefix={arXiv},
    primaryClass={cs.LG},
    url={https://arxiv.org/abs/2402.18846}, 
}

@article{regazzoni2021physics,
    title={A physics-informed multi-fidelity approach for the estimation of differential equations parameters in low-data or large-noise regimes},
    author={Regazzoni, Francesco and Pagani, Stefano and Cosenza, Alessandro and Lombardi, Alessandro and Quarteroni, Alfio},
    journal={Rendiconti Lincei},
    volume={32},
    number={3},
    pages={437--470},
    year={2021},
    doi={10.4171/RLM/943}
}

@article{demo2023deeponet,
    author  = {Demo, Nicola and Tezzele, Marco and Rozza, Gianluigi},
    title   = {A DeepONet Multi-Fidelity Approach for Residual Learning in Reduced Order Modeling},
    journal = {Advanced Modeling and Simulation in Engineering Sciences},
    volume  = {10},
    number  = {1},
    pages   = {12},
    year    = {2023},
    doi     = {10.1186/s40323-023-00249-9},
}

@article{HOWARD2023112462,
    title = {Multifidelity deep operator networks for data-driven and physics-informed problems},
    journal = {Journal of Computational Physics},
    volume = {493},
    pages = {112462},
    year = {2023},
    issn = {0021-9991},
    doi = {10.1016/j.jcp.2023.112462},
    author = {Amanda A. Howard and Mauro Perego and George Em Karniadakis and Panos Stinis},
}

@article{godino,
    title = {Review of multi-fidelity models},
    journal = {Advances in Computational Science and Engineering},
    volume = {1},
    number = {4},
    pages = {351-400},
    year = {2023},
    issn = {},
    doi = {10.3934/acse.2023015},
    author = {M. Giselle Fernández-Godino},
}

@article{piazzola2023comparing,
    author  = {Piazzola, Chiara and Tamellini, Lorenzo and Pellegrini, Riccardo and Broglia, Riccardo and Serani, Andrea and Diez, Matteo},
    title   = {Comparing Multi-Index Stochastic Collocation and Multi-Fidelity Stochastic Radial Basis Functions for Forward Uncertainty Quantification of Ship Resistance},
    journal = {Engineering with Computers},
    volume  = {39},
    number  = {3},
    pages   = {2209--2237},
    year    = {2023},
    doi     = {10.1007/s00366-021-01588-0},
}

@article{KENT2026114761,
    title = {Noise-robust multi-fidelity surrogate modelling for parametric partial differential equations},
    journal = {Journal of Computational Physics},
    volume = {554},
    pages = {114761},
    year = {2026},
    issn = {0021-9991},
    doi = {10.1016/j.jcp.2026.114761},
    author = {Benjamin M. Kent and Lorenzo Tamellini and Matteo Giacomini and Antonio Huerta},
}

@article{Peridkaris2,
    author = {Perdikaris, P. and Raissi, M. and Damianou, A. and Lawrence, N. D. and Karniadakis, G. E.},
    title = {Nonlinear information fusion algorithms for data-efficient multi-fidelity modelling},
    journal = {Proceedings of the Royal Society A: Mathematical, Physical and Engineering Sciences},
    volume = {473},
    number = {2198},
    pages = {20160751},
    year = {2017},
    month = {02},
    doi = {10.1098/rspa.2016.0751},
}

@article{davis,
    doi = {10.1007/s10543-025-01058-9},
    title={Residual multi-fidelity neural network computing},
    author={Davis, Owen and Motamed, Mohammad and Tempone, Raul},
    journal={BIT Numerical Mathematics},
    volume={65},
    number={2},
    pages={15},
    year={2025},
    publisher={Springer}
}

@article{GUO2022114378,
    title = {Multi-fidelity regression using artificial neural networks: Efficient approximation of parameter-dependent output quantities},
    journal = {Computer Methods in Applied Mechanics and Engineering},
    volume = {389},
    pages = {114378},
    year = {2022},
    issn = {0045-7825},
    doi = {10.1016/j.cma.2021.114378},
    author = {Mengwu Guo and Andrea Manzoni and Maurice Amendt and Paolo Conti and Jan S. Hesthaven},
}

@article{e22091022,
    title={Multi-fidelity aerodynamic data fusion with a deep neural network modeling method},
    doi = {10.3390/e22091022},
    author={He, Lei and Qian, Weiqi and Zhao, Tun and Wang, Qing},
    journal={Entropy},
    volume={22},
    number={9},
    pages={1022},
    year={2020},
    publisher={MDPI}
}

@article{SAJJADINIA2022105699,
    doi = {10.1016/j.compbiomed.2022.105699},
    title={Multi-fidelity surrogate modeling through hybrid machine learning for biomechanical and finite element analysis of soft tissues},
    author={Sajjadinia, Seyed Shayan and Carpentieri, Bruno and Shriram, Duraisamy and Holzapfel, Gerhard A},
    journal={Computers in Biology and Medicine},
    volume={148},
    pages={105699},
    year={2022},
    publisher={Elsevier}
}

@article{conti_mf,
    author = {Conti, Paolo and Guo, Mengwu and Manzoni, Andrea and Frangi, Attilio and Brunton, Steven L. and Nathan Kutz, J.},
    title = {Multi-fidelity reduced-order surrogate modelling},
    journal = {Proceedings of the Royal Society A: Mathematical, Physical and Engineering Sciences},
    volume = {480},
    number = {2283},
    pages = {20230655},
    year = {2024},
    month = {02},
    issn = {1364-5021},
    doi = {10.1098/rspa.2023.0655},
}

@article{MENG2025113651,
    title = {Sparse discovery of differential equations based on multi-fidelity Gaussian process},
    journal = {Journal of Computational Physics},
    volume = {523},
    pages = {113651},
    year = {2025},
    issn = {0021-9991},
    doi = {10.1016/j.jcp.2024.113651},
    author = {Yuhuang Meng and Yue Qiu},
}

@article{ruppert,
    author = {David Ruppert and M. P. Wand and Ulla Holst and Ola HöSJER},
    title = {Local Polynomial Variance-Function Estimation},
    journal = {Technometrics},
    volume = {39},
    number = {3},
    pages = {262--273},
    year = {1997},
    publisher = {Taylor \& Francis},
    doi = {10.1080/00401706.1997.10485117},
}

@article{Steinier1964SmoothingAD,
  title={Smoothing and differentiation of data by simplified least squares procedures},
  author={Savitzky, Abraham and Golay, Marcel JE},
  journal={Analytical chemistry},
  volume={36},
  number={8},
  pages={1627--1639},
  year={1964},
  publisher={ACS Publications},
  doi={10.1021/ac60214a047}
}

@article{SEELINGER2025113542,
title = {Democratizing uncertainty quantification},
journal = {Journal of Computational Physics},
volume = {521},
pages = {113542},
year = {2025},
doi = {10.1016/j.jcp.2024.113542},
author = {Linus Seelinger and Anne Reinarz and Mikkel B. Lykkegaard and Robert Akers and Amal M.A. Alghamdi and David Aristoff and Wolfgang Bangerth and Jean Bénézech and Matteo Diez and Kurt Frey and John D. Jakeman and Jakob S. Jørgensen and Ki-Tae Kim and Benjamin M. Kent and Massimiliano Martinelli and Matthew Parno and Riccardo Pellegrini and Noemi Petra and Nicolai A.B. Riis and Katherine Rosenfeld and Andrea Serani and Lorenzo Tamellini and Umberto Villa and Tim J. Dodwell and Robert Scheichl},
}

@article{ZACCHEI2024104902,
    title = {Neural networks based surrogate modeling for efficient uncertainty quantification and calibration of MEMS accelerometers},
    journal = {International Journal of Non-Linear Mechanics},
    volume = {167},
    pages = {104902},
    year = {2024},
    issn = {0020-7462},
    doi = {10.1016/j.ijnonlinmec.2024.104902},
    author = {Filippo Zacchei and Francesco Rizzini and Gabriele Gattere and Attilio Frangi and Andrea Manzoni},
}

@article{trapping,
    author = {Peng, Mai and Kaptanoglu, Alan A. and Hansen, Christopher J. and Stevens-Haas, Jacob and Manohar, Krithika and Brunton, Steven L.},
    title = {Extending the trapping theorem to provide local  stability guarantees for quadratically nonlinear models},
    journal = {Physics of Fluids},
    volume = {37},
    number = {10},
    pages = {107115},
    year = {2025},
    month = {10},
    doi = {10.1063/5.0287432},
}

@article{colbrook,
    author = {Bou-Sakr-El-Tayar, Maria and Bramburger, Jason J. and Colbrook, Matthew},
    title = {Weighted Birkhoff averages accelerate data-driven methods},
    journal = {Proceedings of the Royal Society A: Mathematical, Physical and Engineering Sciences},
    volume = {482},
    number = {2333},
    pages = {20250979},
    year = {2026},
    month = {03},
    issn = {1364-5021},
    doi = {10.1098/rspa.2025.0979},
}

@article{zacchei2026multi,
  title={Multi-fidelity delayed acceptance: Hierarchical MCMC sampling for Bayesian inverse problems combining multiple solvers through deep neural networks},
  author={Zacchei, Filippo and Conti, Paolo and Frangi, Attilio and Manzoni, Andrea},
  journal={Computer Methods in Applied Mechanics and Engineering},
  volume={456},
  pages={118916},
  year={2026},
  doi={10.1016/j.cma.2026.118916}
}

@misc{willard2022integratingscientificknowledgemachine,
      title={Integrating Scientific Knowledge with Machine Learning for Engineering and Environmental Systems}, 
      author={Jared Willard and Xiaowei Jia and Shaoming Xu and Michael Steinbach and Vipin Kumar},
      year={2022},
      eprint={2003.04919},
      archivePrefix={arXiv},
      primaryClass={physics.comp-ph},
      url={https://arxiv.org/abs/2003.04919}, 
}

@misc{raissi2016deepmultifidelitygaussianprocesses,
      title={Deep Multi-fidelity Gaussian Processes}, 
      author={Maziar Raissi and George Karniadakis},
      year={2016},
      eprint={1604.07484},
      archivePrefix={arXiv},
      primaryClass={cs.LG},
      url={https://arxiv.org/abs/1604.07484}, 
}

@misc{villatoro,
      title={Assessing the performance of correlation-based multi-fidelity neural emulators}, 
      author={Cristian J. Villatoro and Gianluca Geraci and Daniele E. Schiavazzi},
      year={2025},
      eprint={2512.02868},
      archivePrefix={arXiv},
      primaryClass={cs.LG},
      url={https://arxiv.org/abs/2512.02868}, 
}

@article{conti2026progressivemultifidelitylearningneural,
    title = {Progressive multi-fidelity learning with neural networks for physical system predictions},
    journal = {Computer Methods in Applied Mechanics and Engineering},
    volume = {455},
    pages = {118881},
    year = {2026},
    issn = {0045-7825},
    doi = {https://doi.org/10.1016/j.cma.2026.118881},
    author = {Paolo Conti and Mengwu Guo and Attilio Frangi and Andrea Manzoni},
    }

@misc{oxby2024confidenceintervalssavitzkygolayfilter,
      title={Confidence Intervals for the Savitzky-Golay Filter with an Application to the Keeling Data for Atmospheric CO2}, 
      author={Paul W. Oxby},
      year={2024},
      eprint={2412.15458},
      archivePrefix={arXiv},
      primaryClass={eess.SP},
      url={https://arxiv.org/abs/2412.15458}, 
}

@book{gardiner2009stochastic,
  title={Stochastic Methods: A Handbook for the Natural and Social Sciences},
  author={Gardiner, C.},
  isbn={9783540707127},
  lccn={2008936877},
  series={Springer Series in Synergetics},
  url={https://books.google.it/books?id=otg3PQAACAAJ},
  year={2009},
  publisher={Springer Berlin Heidelberg}
}

@article{rowbottom2025multi,
title = {Multi-Level Monte Carlo training of neural operators},
journal = {Computer Methods in Applied Mechanics and Engineering},
volume = {453},
pages = {118800},
year = {2026},
doi = {10.1016/j.cma.2026.118800},
author = {James Rowbottom and Stefania Fresca and Pietro Lio and Carola-Bibiane Schönlieb and Nicolas Boullé},
}

@article{qian2024multifidelity,
    author = {Qian, Elizabeth and Kang, Dayoung and Sella, Vignesh and Chaudhuri, Anirban},
    year = {2025},
    month = {01},
    pages = {271-297},
    title = {Multifidelity linear regression for scientific machine learning from scarce data},
    volume = {7},
    journal = {Foundations of Data Science},
    doi = {10.3934/fods.2024049}
}
\end{document}